\documentclass[10pt,twocolumn,letterpaper]{article}

\usepackage{cvpr}      % To produce the REVIEW version
\definecolor{cvprblue}{rgb}{0.21,0.49,0.74}
\usepackage[pagebackref,breaklinks,colorlinks,allcolors=cvprblue]{hyperref}
\usepackage[hyphens]{url}  % DO NOT CHANGE THIS
\usepackage{graphicx} % DO NOT CHANGE THIS
\usepackage{natbib}  % DO NOT CHANGE THIS AND DO NOT ADD ANY OPTIONS TO IT
\usepackage{caption} % DO NOT CHANGE THIS AND DO NOT ADD ANY OPTIONS TO IT
\usepackage{algorithm}
\usepackage{algorithmic}

\usepackage{newfloat}
\usepackage{listings}
\DeclareCaptionStyle{ruled}{labelfont=normalfont,labelsep=colon,strut=off} % DO NOT CHANGE THIS
\floatstyle{ruled}
\newfloat{listing}{tb}{lst}{}
\floatname{listing}{Listing}

\usepackage{booktabs}

\usepackage{amsmath}
\usepackage{amssymb}
\usepackage{subcaption}
\usepackage{booktabs}
\usepackage{multirow}

\newcommand{\best}[1]{\textcolor{red}{\textbf{#1}}}
\newcommand{\secondbest}[1]{\textcolor{blue}{\underline{#1}}}
\newcommand{\thirdbest}[1]{\textcolor{teal}{\textit{#1}}}

\title{Bridging Restoration and Generation in One-step Diffusion for Real-World Image Super-Resolution}
\author{
    Shyang-En Weng$^{1}$ \qquad
    Yi-Cheng Liao$^{1}$ \qquad
    Yu-Syuan Xu$^{2}$ \qquad
    Chia-Hung Yuan$^{2}$ \\
    Wei-Chen Chiu$^{1}$ \qquad
    Ching-Chun Huang$^{1}$\thanks{Corresponding author (chingchun@nycu.edu.tw)} \\
    $^{1}$Department of Computer Science, National Yang Ming Chiao Tung University, Taiwan \\
    $^{2}$MediaTek Inc., Taiwan
}

\begin{document}

\maketitle

\begin{abstract}
    Pretrained diffusion models have revolutionized real-world image super-resolution (Real-ISR), but their iterative sampling is computationally prohibitive, driving efforts to distill it into a single step.
    General one-step methods fine-tune the generative prior into a deterministic mapping, restoring efficiency but discarding its stochastic nature.
    Conversely, recent attempts re-engage generation by shifting the timestep or injecting random noise, adjusting either the position or the state while the other stays fixed. Because only one side is controlled, the two align at isolated preset timesteps but drift apart once steered, leaving generation unstable.
    To address this, we present one-step diffusion via Inversion and Degradation-aware Sampling for Real-ISR (IDaS-SR), a one-step framework that bridges deterministic restoration and stochastic generation. 
    At its core, Manifold Anchoring grounds the low-quality latent on the pretrained trajectory through two operations jointly estimated by the Manifold Inversion Noise Estimator (MINE): positioning declares where the latent lies and how it deviates from the clean state, while inversion aligns the latent to the declared position.
    Upon the anchor, CHARIOT reintroduces controlled stochasticity by jointly rescheduling the trajectory and interpolating the noise, enabling a single scalar to smoothly navigate the fidelity-realism trade-off. 
    Extensive experiments demonstrate that IDaS-SR effectively unleashes the generative prior, achieving state-of-the-art performance under explicit control in a single inference step.
\end{abstract}

\section{Introduction}
\label{sec:intro}

% A. Real-ISR intro
Real-world Image Super-Resolution (Real-ISR) is a classic and foundational challenge in low-level vision tasks that aims to restore high-quality (HR) images from low-quality (LQ) inputs. Due to complex, unknown, and diverse degradation mechanisms, it becomes a highly ill-posed inverse problem that necessitates enhancing fidelity and realism to align with human visual perception.

% B. T2I diffusion and multi-step diffusion SR
% C. The deterministic one-step restoration limited the generative power of its through two perspectives: timestep settings and LQ mismatch
To address this, recent research leverages the rich prior knowledge encapsulated in generative models. Pretrained text-to-image (T2I) diffusion models, such as Stable Diffusion (SD), have shown strong potential for generating high-quality natural synthesis and serve as powerful generative priors for Real-ISR. 
Multi-step diffusion models such as \cite{stablesr, diffbir, pasd, seesr} achieve superior perceptual quality by iteratively refining through a learned denoising process, effectively leveraging a generalized prior to ensure realistic and faithful restorations. 
Although these methods offer strong generative capabilities, multi-step denoising during inference incurs high computational costs and significant latency.

\begin{figure*}[tb]
    \centering
    \resizebox{.8\linewidth}{!}{
    % \begin{subfigure}{0.49\linewidth}
    %     \includegraphics[width=\linewidth]{Fig/traj/deg-aware OSEDiff-Restoration orientation.pdf}
    %     \caption{}
    % \end{subfigure}
    % \hfill
    % \begin{subfigure}{0.49\linewidth}
    %     \includegraphics[width=\linewidth]{Fig/traj/deg-aware OSEDiff-Diffusion orientation.pdf}
    %     \caption{}
    % \end{subfigure}
    % \begin{subfigure}{0.45\linewidth}
    %     \includegraphics[width=\linewidth]{Fig/traj/deg-aware OSEDiff-Diffusion-Restoration orientation.jpg}
    %     \caption{Another example of a subfigure}
    %     \label{fig:short-b}
    % \end{subfigure}
    % \hfill
    % \vfill
    % \begin{subfigure}{0.32\linewidth}
    %     \includegraphics[width=\linewidth]{Fig/traj/deg-aware OSEDiff-Extended Diffusion-Restoration orientation.pdf}
    %     \caption{}
    % \end{subfigure}
    
    \begin{subfigure}{.25\linewidth}
        \includegraphics[width=\linewidth]{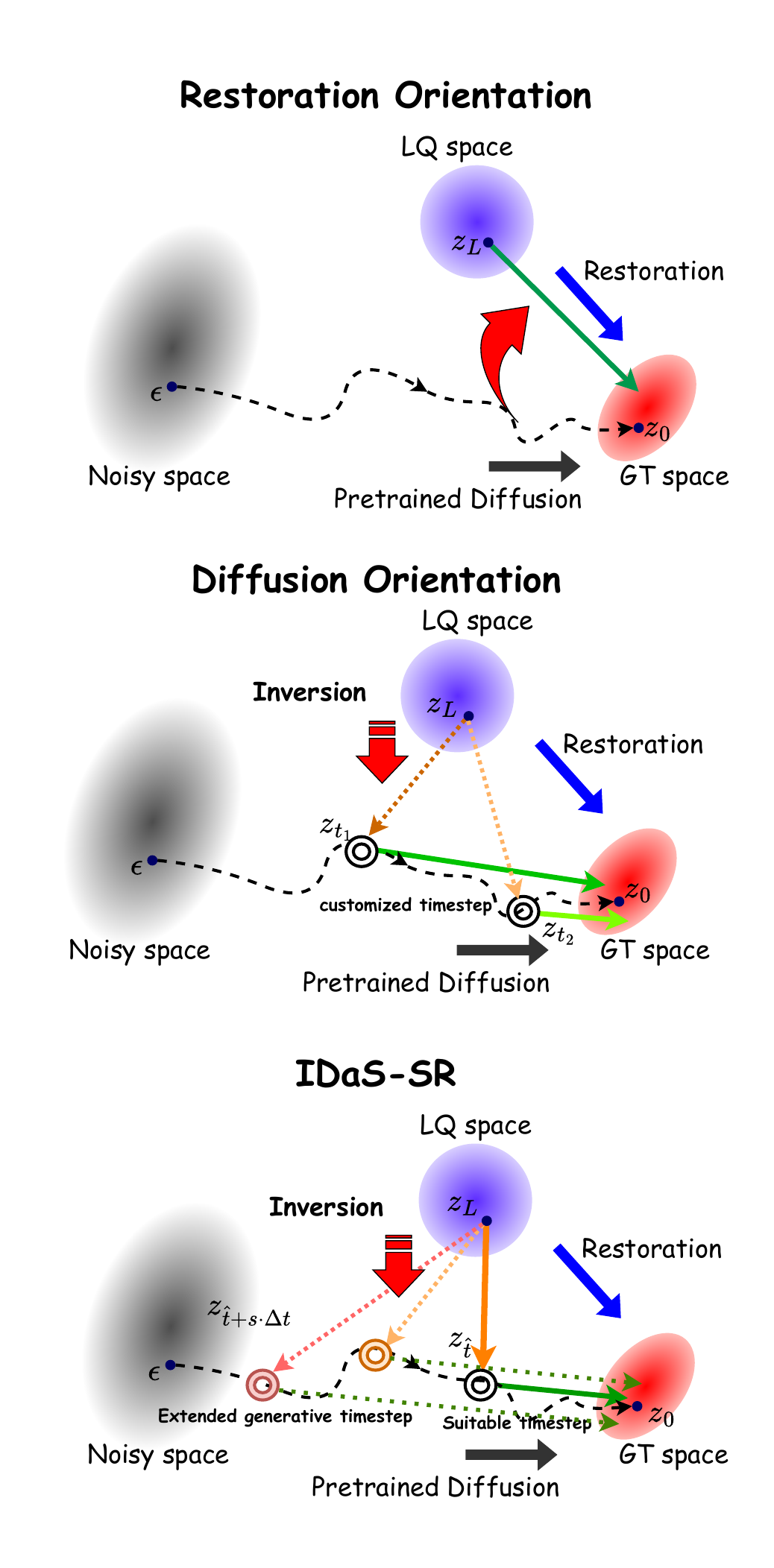}
        \caption{}
    \end{subfigure}
    \hfill
    \begin{subfigure}{.54\linewidth}
        \includegraphics[width=\linewidth]{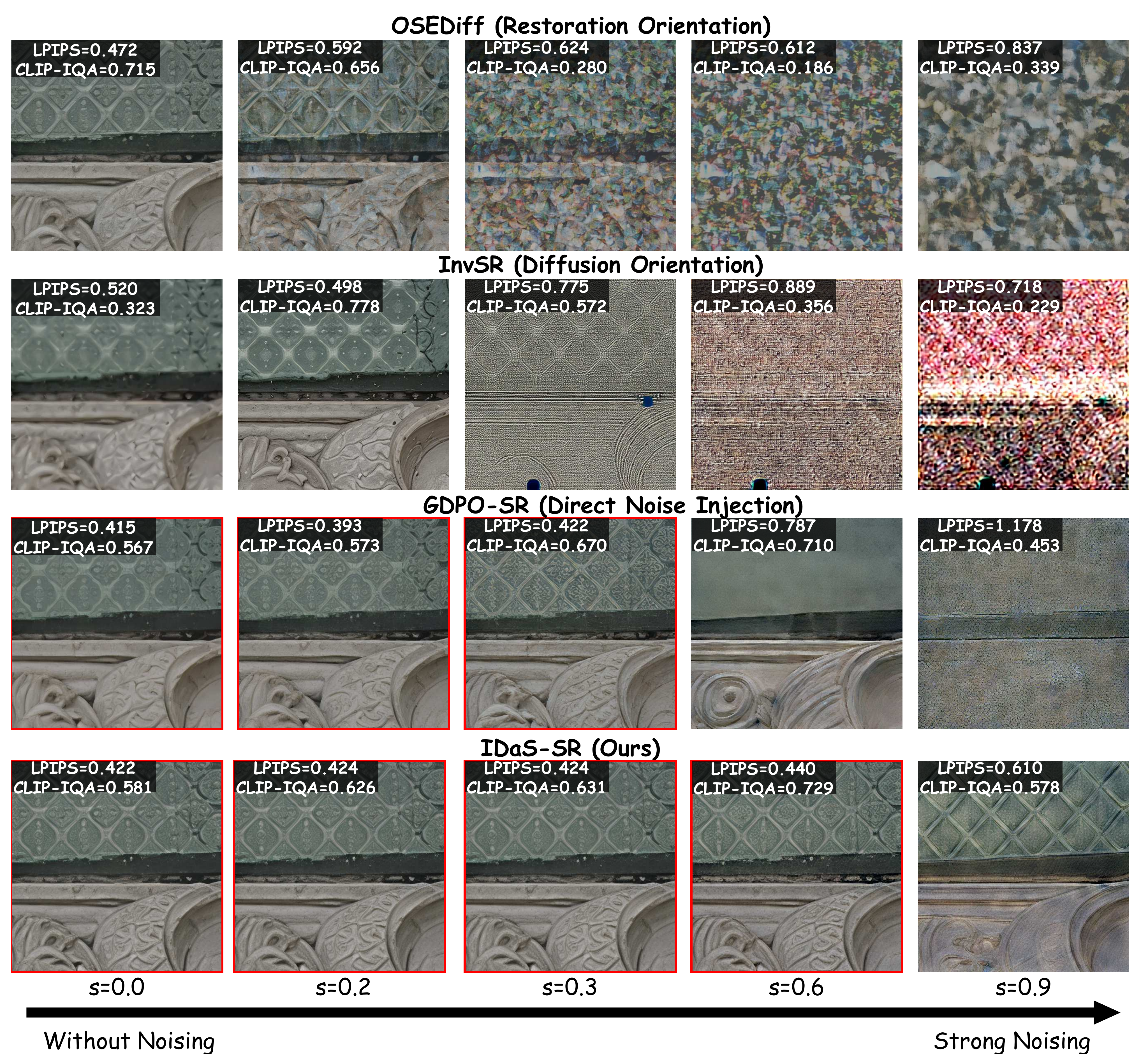}
        \caption{}
    \end{subfigure}
    }
    \caption{Challenges in steering one-step diffusion for Real-ISR.
        (a) Entering the diffusion trajectory requires a timestep and a matching noise term, yet existing methods determine only one: \emph{Restoration Orientation} feeds the latent unchanged at an extreme timestep, \emph{Diffusion Orientation} enters at a timestep chosen beforehand, and \emph{Direct Noise Injection} perturbs the latent while the timestep stays fixed. IDaS-SR determines both.
        (b) Steering strength increases left to right, given by $s$ for IDaS-SR and by the approximately corresponding noising levels $t=\{50, 250, 450, 700, 900\}$ for the baselines, applied through each method's own control where available. Existing methods lose structure once pushed beyond their calibrated settings, whereas IDaS-SR retains it at its default $s{=}0.6$. Red outlines mark structurally faithful results.
    % Conceptual comparison of manifold mapping trajectories. 
    % (a) Conventional one-step diffusion restoration~\cite{osediff, pisasr, tsdsr}, which enforces a direct, rigid mapping from LQ space to the ground-truth manifold. (b) Adaptive inversion~\cite{invsr} utilizing customized timesteps to better align the restoration path with the diffusion trajectory. (c) Our proposed IDaS-SR framework, which synergizes adaptive manifold mapping to establish a deterministic restoration anchor (solid line), and with extended generative rescheduling (dotted line) to facilitate controlled texture synthesis.
    }
  \label{fig:manifold_diagram}
\end{figure*}

% One-step SR & issues
To reconcile the generative priors of diffusion models with inference efficiency, recent research distills multi-step models into single-step mappings.
\emph{Restoration Orientation methods}~\cite{osediff, pisasr, tsdsr} reframe T2I diffusion as a deterministic LQ-to-HQ mapping by fine-tuning the VAE encoder and the denoising backbone, aligning the generative prior with the restoration objective.
Compressing the trajectory into a single forward pass at a fixed timestep, however, reduces the versatile prior to a rigid mapping (Fig.~\ref{fig:manifold_diagram}(a), top).

% Anchoring in plain terms + two lines
Re-engaging generation requires placing the LQ latent back onto the pretrained trajectory, an operation we refer to as \emph{anchoring}: the timestep specifies \emph{the position}, while the latent and the \emph{inversion noise} that carries it there compose \emph{the state} that position implies.
Pretraining always pairs the two, corrupting the latent exactly to the level its timestep announces, yet existing one-step methods break this pairing in different ways (Fig.~\ref{fig:manifold_diagram}(a), middle).
% \emph{Restoration Orientation}~\cite{osediff, pisasr, tsdsr} declares an extreme position and feeds $\mathbf{z}_{L}$ as the state, absorbing the disagreement by fine-tuning until the prior degenerates into a deterministic mapping.
\emph{Diffusion Orientation}~\cite{invsr, omgsr} moves the position inward to reduce the disagreement, computing where to enter without reconciling the state~\cite{omgsr}, or learning a state that fits only a few steps fixed beforehand~\cite{invsr}.
\emph{Direct Noise Injection}~\cite{upsr, hypir, gdposr} perturbs the state with injected noise, so as the injected level rises the state drifts from the position that describes it.
% GDPO-SR~\cite{gdposr} makes the split explicit, injecting noise at one timestep while denoising at another and exposing the injection level as a user control; as that control is swept, the denoising timestep stays where it was, so the state drifts from the position that describes it.
% In each case the declared position and the received state never move together, so pushing toward generation leaves the prior with an input it was never trained on, and structure degrades into artifacts (Fig.~\ref{fig:manifold_diagram}(b)).
In each case, the declared position and received state are misaligned; pushing toward generation exposes the prior to out-of-distribution inputs, leading to structural degradation and artifacts (Fig.~\ref{fig:manifold_diagram}(b)).

% While recent attempts aim to address these initialization issues, they often lack sufficient adaptability: predictive approaches like \cite{herosr} can suffer from optimization instability during timestep estimation, while \cite{rcod, omgsr} rely on static training set statistics, which often fail to account for instance-specific degradation levels, limiting their adaptability to diverse real-world scenarios.

% Ours + Contributions
We instead specify both sides of the anchor and make it traversable: since the anchor is what keeps the traversal stable, the two are inseparable rather than merely complementary (Fig.~\ref{fig:manifold_diagram}(a), bottom).
Our main contributions are summarized as follows:
\begin{itemize}
    \item We present Inversion and Degradation-aware Sampling for Real-ISR (IDaS-SR), an \emph{anchor-and-steer} one-step framework that recasts Real-ISR as a controlled traversal of the pretrained diffusion manifold, in which a single model acts as both a deterministic restorer and a perceptual generator within one forward pass.
    \item We propose Manifold Anchoring, realized by the Manifold Inversion Noise Estimator (MINE), which grounds the generative prior into a deterministic restoration anchor. \emph{Positioning} declares where the latent enters through an anchoring timestep $\hat{t}$, estimated under the restoration objective and paired with a degradation context $\hat{c}_{\mathrm{deg}}$ that bridges the LQ observation to its HQ counterpart; \emph{inversion} then estimates the noise map $\hat{\varepsilon}_{\mathrm{inv}}$ that absorbs its corruption, so the state the prior receives is the one its position announces.
    \item We devise Controlled Hallucination via Anchored Rescheduling of Inversion Over Trajectories (CHARIOT), which reopens this restoration anchor toward generation. 
    Governed by a single scalar $s$, CHARIOT raises the declared timestep and interpolates $\hat{\varepsilon}_{\mathrm{inv}}$ toward Gaussian noise, driving the anchored state to highly realistic generation while keeping position and state aligned; the same $s$ exposes the fidelity--realism balance as a continuous inference-time control.
\end{itemize}

\section{Related Work}

% A. Real-ISR
% \vspace{0.1in}
\noindent\textbf{Diffusion-based Real-ISR.}
Early deep learning approaches~\cite{srcnn, edsr, srgan, swinir} formulate Real-ISR as deterministic regression and achieve strong PSNR and SSIM, but assume simple degradations such as bicubic downsampling. BSRGAN~\cite{bsrgan} and Real-ESRGAN~\cite{realesrgan} extend them to in-the-wild images through synthetic degradation pipelines and adversarial training, yet GAN-based models remain unstable to train and prone to structural artifacts under severe degradations. Diffusion models~\cite{ddpm} have since become the dominant paradigm for perceptual quality in Real-ISR: leveraging pretrained text-to-image (T2I) priors~\cite{ldm,sdxl}, they synthesize high-fidelity textures unattainable through pixel-level optimization~\cite{stablesr, pasd, diffbir, resshift, supir, seesr, faithdiff}. A central question is how to condition the generative process on the low-quality (LQ) input to preserve structure: DiffBIR~\cite{diffbir} adopts a two-stage restore-then-generate pipeline, PASD~\cite{pasd} introduces pixel-aware cross-attention for structure-aligned guidance, SeeSR~\cite{seesr} adds a degradation-aware prompt encoder to balance semantic correctness with fine detail, and FaithDiff~\cite{faithdiff} aligns LQ features with the diffusion manifold at the latent level. Despite their perceptual quality, the iterative sampling of these multi-step models incurs high inference latency, precluding the real-time deployment afforded by GAN-based methods.

% B. Diffusion timestep as degradation intensity
% \vspace{0.1in}
\noindent\textbf{One-step Diffusion for Real-ISR.}
To mitigate multi-step diffusion latency, recent research focuses on distilling generative priors into efficient one-step architectures. 
OSEDiff~\cite{osediff} and SinSR~\cite{sinsr} utilize low-quality (LQ) initializations and consistency-preserving distillation to condense the reverse trajectory, while PiSA-SR~\cite{pisasr} employs latent residual learning to decouple structural fidelity from semantic synthesis. 
Other strategies emphasize trajectory calibration: TSD-SR~\cite{tsdsr} leverages ground-truth references via Target Score Distillation to refine detail recovery, and FluxSR~\cite{fluxsr} relates the LR-to-HR restoration flow to the frozen noise-to-image generation flow via flow trajectory distillation, though the learned mapping is fixed once training ends. 
Despite these advances, existing methods reduce the generative prior to a deterministic mapping, discarding its stochastic nature.
% Our work addresses this by reformulating the connection between generation and restoration, elegantly bridging both manifolds to unleash generative potential without sacrificing structural integrity.

% \vspace{0.1in}
\noindent\textbf{Generative Flexibility in Diffusion-based Real-ISR.}
Since the general one-step design restricts generation from the outset, recent work recovers generative capacity by revisiting the starting state along the two remaining lines introduced in Sec.~\ref{sec:intro}.
\textit{Diffusion Orientation} moves the entry inward to an intermediate position.
OMGSR~\cite{omgsr} retains the direct-injection design and pre-computes a single SNR-matched timestep offline, revising the position while the latent it receives stays unchanged.
InvSR~\cite{invsr} instead learns an inversion network that predicts the noise realizing a valid state at the entry step, yet trains it only on a handful of manually selected steps~\cite{omgsr}, leaving no trained path between them.
\textit{Direct Noise Injection} keeps the position and perturbs the state.
UPSR~\cite{upsr} weights the injected noise by per-pixel uncertainty, sparing flat regions to preserve LQ information while perturbing textured ones for synthesis.
HYPIR~\cite{hypir} describes an injection-scale control at inference, reported for its larger variants, and GDPO-SR~\cite{gdposr} makes the split explicit, injecting noise at one timestep and denoising at another, yet sweeping only the former so the denoising timestep stays where it was.
In essence, each realizes at most half of an anchor: the position is revised without a matching state, the state is matched only at preset positions, or the state is perturbed while the position stays behind.
Consistency therefore survives only at isolated points, beyond which the state no longer matches what the position announces.
IDaS-SR addresses this by specifying both sides of the anchor (Manifold Anchoring) and advancing them in concert (CHARIOT), maintaining consistency across the full restoration--generation spectrum.

\section{Methodology}
\subsection{Preliminary}
\noindent\textbf{Problem Formulation.} 
The real-world image super-resolution (Real-ISR) task aims to reconstruct a high-quality (HQ) image $x_{H}$ from a low-quality (LQ) input $x_{L}$ that has suffered from complex, unknown degradations. Formally, training a parameterized restoration model $G_{\theta}$ can be modeled as minimizing the following objective:
\begin{equation}
\label{eq:restore_optim}
\resizebox{0.95\columnwidth}{!}{$
    \theta^{*} = \arg\min_{\theta} \mathbb{E}_{(x_{L}, x_{H})} \left[ \mathcal{L}_{data}(G_{\theta}(x_{L}), x_{H}) + \lambda \mathcal{L}_{reg}(G_{\theta}(x_{L})) \right],
$}
\end{equation}
where $\mathcal{L}_{data}$ represents the data fidelity term (e.g., $L_2$ or LPIPS) measuring the discrepancy between the predicted image $\hat{x}_{H} = G_{\theta}(x_{L})$ and the ground truth $x_{H}$. The regularization term $\mathcal{L}_{reg}$ (e.g., GAN, VSD) is crucial for integrating natural image priors, fundamentally acting to align the distribution of the generated images $q_{\theta}(\hat{x}_{H})$ with the true natural image distribution $p(x_{H})$ by minimizing their KL-divergence: $\mathcal{D}_{KL}(q_{\theta}(\hat{x}_{H}) || p(x_{H}))$. 

\vspace{0.1in}
\noindent\textbf{Latent Diffusion Models.} 
Generative diffusion models provide powerful priors for image generation by operating in the compressed latent space of a pretrained autoencoder. During the forward process, the model diffuses an input latent feature $z$ through:
\begin{equation}
    z_{t} = \alpha_{t} z_0 + \beta_{t} \varepsilon, \quad \varepsilon \sim \mathcal{N}(0, \mathbf{I}),
    \label{eq.diff_forward}
\end{equation}    
where $\alpha_{t}$ are noise schedule scalars at timestep $t \in \{1, \dots, T\}$, and $\varepsilon$ represents Gaussian noise. In the reverse phase, a denoising network predicts the added noise $\hat{\varepsilon} = \varepsilon_{\theta}(z_{t}; t, c_{y})$ given text conditioning $c_{y}$, allowing the denoised latent to be estimated as:
\begin{equation}
    \hat{z}_{0} = \frac{z_{t} - \beta_{t} \varepsilon_{\theta}(z_{t}; t, c_{y})}{\alpha_{t}}.
    \label{eq.diff_denoising}
\end{equation}

\vspace{0.1in}
\noindent\textbf{One-Step Diffusion.} 
While traditional multi-step diffusion requires dozens of iterative denoising steps that increase computational cost, the process has usually been condensed. Since the LQ image contains rich structural information, we utilize its latent representation $z_{L} = \mathcal{E}(x_{L})$ from VAE encoder $\mathcal{E}$ directly as the starting point. Formally, the restoration function $F_\theta$ is defined as:
\begin{equation}
    \hat{z}_{H} = F_\theta(z_{L}; c_{y}) \triangleq \frac{z_{L} - \beta_{T} \varepsilon_{\theta}(z_{L}; T, c_{y})}{\alpha_{T}},
    \label{eq.1step_denoising}
\end{equation}
where $\alpha_T$ and $\beta_T$ are time-dependent coefficients following the noise schedule. The reverse process employs a denoising network $\varepsilon_\theta(z_L; T, c_y)$ to predict the noise given timestep $T$ and text conditioning $c_y$.

\begin{figure*}[t]
    \centering
    \includegraphics[width=\linewidth]{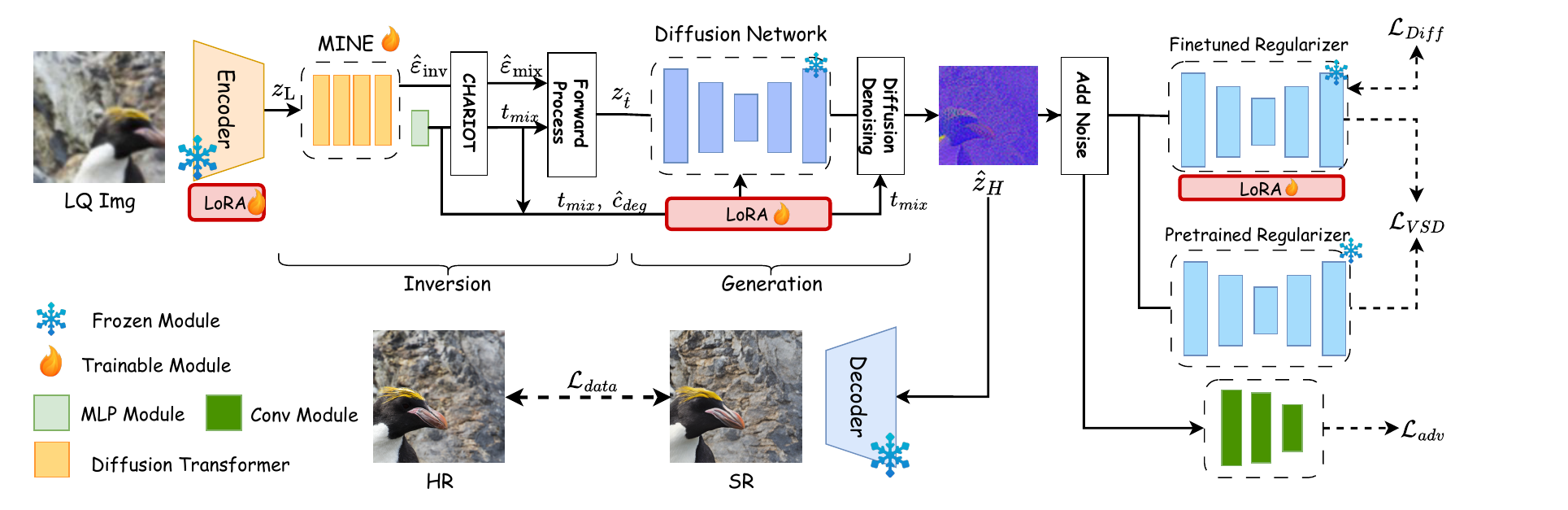}
    \caption{Overview of IDaS-SR. Our framework bridges restoration and generation manifolds. MINE first predicts the positioning ($\hat{t}$ and $\hat{c}_\text{deg}$) and inversion noise ($\hat{\varepsilon}_{\text{inv}}$) to anchor the LQ input. CHARIOT then utilizes a control scalar ($s$) to interpolate between anchored and noise, balancing structural fidelity and perceptual realism.}
    \label{fig:idas-sr}
\end{figure*}

\subsection{Bridging Generative and Restoration for Real-ISR}
% Motivation and High-Level Concept
Both generation and restoration can be formulated through denoising, yet generation seeks distributional realism while restoration demands faithful reconstruction, and each constrains the other.
Exploiting the pretrained generative prior for restoration is nontrivial. The LQ latent does not lie on the diffusion manifold, while enforcing a direct restoration mapping instead restricts the model's inherent generative capabilities.
From another view, the restoration objective also limits the generative capacity. The information bottleneck of the LQ input prevents the recovery of irreversibly lost high-frequency details, regardless of how faithfully the restoration is anchored. Such details can only be synthesized by the generative prior, whose capacity resides in the high-noise regime of the diffusion trajectory.

Inspired by these explorations, we propose Inversion and Degradation-aware Sampling for Real-ISR (IDaS-SR), which unifies both tasks into an effective one-step diffusion paradigm: anchoring the generative process to the deterministic nature of restoration via Manifold Inversion Noise Estimator (MINE) and, in turn, steering restoration back toward controllable generation through a mechanism called Controlled Hallucination via Anchored Rescheduling of Inversion Over Trajectories (CHARIOT). The overall pipeline of IDaS-SR is shown in Fig.~\ref{fig:idas-sr}.

\subsection{Taming Generative Diffusion into Restoration}
Enforcing the diffusion model to learn a direct restoration mapping introduces a distribution mismatch, necessitating anchoring the generation trajectory within a suitable region for restoration.
To address this, we propose MINE, which unifies diffusion inversion and positioning for restoration.
Through MINE, we reframe the restoration challenge: identifying the gap between the LQ latent and the diffusion trajectory, and bridging this gap to seamlessly map the latent into the high-quality (HQ) space $\hat{z}_H$. 
% , and the manifold anchoring concept is shown in Fig.~\ref{fig:manifold_diagram}.

% To retain these capabilities, recent methods instead steer the pretrained generative process toward restoration by supplying auxiliary guidance, e.g., detailed textual prompts describing the image content~\cite{osediff}, semantic feature conditioning~\cite{vosr}, or a suitably initialized timestep~\cite{invsr, herosr, omgsr}. 

% Restoration and generation share the same denoising process, yet a fundamental discrepancy remains: the LQ latent does not lie on the pretrained diffusion manifold. Prior works bridge this gap by observing that the LQ latent resembles an intermediate noisy state on the diffusion trajectory, either scheduling a matched timestep for latent injection~\cite{invsr, herosr, omgsr} or injecting auxiliary noise to align the latent with the noisy distribution~\cite{upsr, hypir}.
% However, both remedies rely on fixed or dataset-level heuristics that neglect the instance-specific degradation of each input, inevitably leading to under-restoration for severe degradations or over-denoising for mild ones. To this end, we introduce the \textbf{Manifold Inversion Noise Estimator} (MINE), a severity-aware estimator that dynamically projects each LQ latent onto its optimal anchor on the diffusion manifold, jointly resolving timestep initialization and distributional trajectory mismatch.

\vspace{0.1in} 
\noindent\textbf{Manifold Inversion Noise Estimator (MINE).}
Given a low-quality (LQ) latent $z_L$, MINE is formulated as $(\hat{\varepsilon}_{\text{inv}}, \bar{p}) {=} f_{\phi}(z_{L})$, where the inversion noise $\hat{\varepsilon}_{\text{inv}}$ realizes the state, and the pooled positioning context $\bar{p}$ is routed through lightweight MLP heads into the positioning pair $(\hat{t}, \hat{c}_{\text{deg}}) {=} \text{MLP}(\bar{p})$, comprising the anchoring timestep $\hat{t}$ and the degradation embedding $\hat{c}_{\text{deg}}$.
This two-branch design realizes manifold anchoring by decoupling inversion from positioning.
The inversion noise map $\hat{\varepsilon}_{\text{inv}}$ absorbs the corruption carried by the LQ observation into the noise term, projecting it onto the pretrained trajectory, while $\hat{t}$ locates the anchor and $\hat{c}_{\text{deg}}$ provides degradation-aware conditioning of our framework, bridging the LQ to its HQ counterpart for the frozen backbone.
The anchored latent is then constructed as:
\begin{equation}
    z_{\hat{t}} = \alpha_{\hat{t}} z_{L} + \beta_{\hat{t}}\, \hat{\varepsilon}_{\text{inv}},
    \label{eq:idas_inversion}
\end{equation}
where $\alpha_{\hat{t}} {=} \sqrt{\bar{\alpha}_{\hat{t}}}$ and $\beta_{\hat{t}} {=} \sqrt{1-\bar{\alpha}_{\hat{t}}}$ follow the pretrained noise schedule.

\vspace{0.1in}
\noindent\textbf{Anchored One-step Denoising.}
Once the latent is placed on the diffusion trajectory as $z_{\hat{t}}$, the clean latent $\hat{z}_{H}$ is recovered by a single denoising step, rewriting Eq.~\eqref{eq.1step_denoising} as:
\begin{equation}
    \hat{z}_{H} \triangleq \frac{z_{\hat{t}} - \beta_{\hat{t}}\, \varepsilon_{\theta}(z_{\hat{t}};\, \hat{t},\, c_{y} + \hat{c}_{\text{deg}})}{\alpha_{\hat{t}}},
    \label{eq:idas_denoising}
\end{equation}
where $\varepsilon_{\theta}$ denotes the LoRA-adapted diffusion network and $c_{y}$ the semantic text conditioning.
The two terms act through different interfaces: $\hat{t}$ enters the timestep embedding to fix where the state lies on the trajectory, while $\hat{c}_{\text{deg}}$ is added to $c_{y}$ and passes through cross-attention to carry the LQ-to-HQ relation that a scalar cannot express.
The positioning pair and the inversion noise together complete the anchor, and the resulting restoration is deterministic while still drawn from the generative prior.

% \noindent\textbf{Distribution MAtching.}
% Moreover, since the one-step diffusion is geometrically matched to the restoration, the 

% \begin{figure}[t]
%     \centering
%     % Subfigure (a): Wide (0.68)
%     \begin{subfigure}[b]{0.48\linewidth}
%         \centering
%         \includegraphics[width=\linewidth]{Fig/diff-rest_orient_traj.jpg} 
%         \caption{}
%     \end{subfigure}
%     \hfill % Maximizes space between them (approx 4% gap)
%     % Subfigure (b): Narrow (0.28)
%     \begin{subfigure}[b]{0.48\linewidth}
%         \centering
%         \includegraphics[width=\linewidth]{Fig/ex_diff-rest_orient_traj.jpg} 
%         \caption{}
%     \end{subfigure}
    
%     \caption{}
%     \label{fig:traj}
% \end{figure}

\subsection{Unleashing Deterministic Restoration back to Generation}
% Due to LQ info loss, need generation to hallucinate that.
%  All because of different objectives
Due to the inherent information loss in LQ images, Real-ISR relies on generation to hallucinate realistic details, with recent one-step methods re-engaging the prior via inversion or noising~\cite{invsr, herosr, omgsr, hypir, upsr}.
However, these single-axis manipulations rest on states never trained for them, compromising faithful restoration once steered.
Therefore, we propose CHARIOT, which sweeps a single continuous scalar to extend the trajectory from the anchor toward the generative regime, while manifold anchoring guarantees that the generation remains faithfully aligned with the restoration objective throughout.

% Enforcing the diffusion model to learn a direct restoration mapping restricts its inherent generative capabilities, while 

\vspace{0.1in}
\noindent\textbf{Joint Rescheduling and Noise Interpolation for Controlled Generation.}
The anchor alone yields a single anchor. To make it traversable, CHARIOT advances the position and the state together under a control scalar $s {\in} [0,1)$: the anchoring timestep is rescheduled toward a generative limit $T_{\max}$,
\begin{equation}
    t_{\text{mix}} = \hat{t} + s\,(T_{\max} - \hat{t}),
    \label{eq:chariot_t}
\end{equation}
while the estimated inversion noise is interpolated into Gaussian noise,
\begin{equation}
    \hat{\varepsilon}_{\text{mix}} = \text{Norm}\big( (1-s)\,\hat{\varepsilon}_{\text{inv}} + s\,\varepsilon \big),
    \label{eq:chariot_eps}
\end{equation}
where $\varepsilon \sim \mathcal{N}(0,\mathbf{I})$ and $\text{Norm}(\cdot)$ restores unit variance, which the mixture would otherwise lose since $(1-s)^2 + s^2 \le 1$, keeping the state consistent with the position $t_{\text{mix}}$ declares.
The steered latent and its one-step reconstruction then follow Eq.~\eqref{eq:idas_inversion} and Eq.~\eqref{eq:idas_denoising} with $(\hat{t}, \hat{\varepsilon}_{\text{inv}})$ replaced by $(t_{\text{mix}}, \hat{\varepsilon}_{\text{mix}})$:
\begin{equation}
    z_{t_{\text{mix}}} = \alpha_{t_{\text{mix}}} z_{L} + \beta_{t_{\text{mix}}}\, \hat{\varepsilon}_{\text{mix}},
    \label{eq:chariot_z}
\end{equation}
\begin{equation}
    \hat{z}_{H} = \frac{z_{t_{\text{mix}}} - \beta_{t_{\text{mix}}}\, \varepsilon_{\theta}(z_{t_{\text{mix}}};\, t_{\text{mix}},\, c_{y} + \hat{c}_{\text{deg}})}{\alpha_{t_{\text{mix}}}}.
    \label{eq:chariot_denoise}
\end{equation}
During training, $s \sim \mathcal{U}(0,1)$, so the anchor is optimized across the whole range rather than at a single operating point; at inference $s$ is set by the user, with $s{=}0.6$ as our realism default.
Because a fraction of $\hat{\varepsilon}_{\text{inv}}$ persists in the noise term, the state retains instance-specific information even at high noise levels, and the position it declares continues to describe it, which is what keeps the traversal stable as $s$ grows.

\subsection{Reconstruction and Regularization Loss}

To train our one-step diffusion model effectively, following Eq.~\eqref{eq:restore_optim},  we employ a joint objective comprising a base reconstruction loss for structural fidelity and a dual generative regularization strategy to ground the output in real-world distributions. 

The reconstruction objective ($\mathcal{L}_{\text{data}}$) ensures pixel-level alignment and basic perceptual similarity using standard $\ell_2$ and LPIPS metrics:
\begin{equation}
\label{eq:data_loss}
    \mathcal{L}_{\text{data}} = \mathcal{L}_{2} + \lambda_{lpips} \mathcal{L}_{\text{LPIPS}},
\end{equation}
where $\lambda_{lpips}$ controls the influence of the perceptual loss. However, relying exclusively on direct reconstruction targets typically yields over-smoothed high-frequency details. 

To enhance realistic textures and enforce global coherence, we introduce generative regularization ($\mathcal{L}_{\text{reg}}$) by combining Variational Score Distillation (VSD)~\cite{osediff} with a latent GAN. VSD extracts natural-image priors using a frozen, pretrained diffusion regularizer and a finetuned regularizer. By minimizing the score discrepancy between the generated latent and the target manifold, VSD enforces coherent global structures. However, as discussed in \cite{tsdsr, pisasr, hypir, fluxsr, gendr}, standard VSD often lacks direct real-world texture grounding and can degrade when distilled from early-stage noisy latents.

To counteract this over-smoothing and restore perceptual sharpness, we introduce a lightweight Latent Patch Discriminator. Rather than a full-resolution pixel-space critic, the discriminator is a shallow convolutional network that operates directly on the VAE latent and emits a grid of patch logits~\cite{patchgan}, trained jointly with the generator to encourage realistic textures at negligible additional cost. The combined generative regularization loss is formulated as:
\begin{equation}
\label{eq:reg_loss}
    \mathcal{L}_{\text{reg}} = \lambda_{vsd} \mathcal{L}_{\text{VSD}} + \lambda_{adv} \mathcal{L}_{\text{GAN}},
\end{equation}
where $\lambda_{vsd}$ and $\lambda_{adv}$ affect the relative influence of the score distillation and adversarial priors, respectively. $\mathcal{L}_{\text{diff}}$ is used exclusively to update the parameters of the finetuned regularizer in VSD to represent the restored distribution.

Finally, the comprehensive training objective for our one-step diffusion model integrates both the data fidelity constraints and the generative priors:
\begin{equation}
\label{eq:total_loss}
    \mathcal{L}_{\text{total}} = \mathcal{L}_{\text{data}} + \mathcal{L}_{\text{reg}}.
\end{equation}

\section{Experiment Results}
\subsection{Experimental Settings}

\begin{table*}[t]
    \centering
    \small
    \setlength{\tabcolsep}{1mm}
    \begin{tabular}{l   cccc   ccccccc   c}
        \toprule
        & \multicolumn{4}{c}{Multi-step Methods} & \multicolumn{7}{c}{One-step Methods} & \multicolumn{1}{c}{Ours} \\
        \cmidrule(lr){2-5} \cmidrule(lr){6-12} \cmidrule(lr){13-13}
        \textbf{Metric} 
        & SeeSR & FaithDiff & UPSR & DiT4SR 
        & OSEDiff & InvSR & PiSA-SR & TSD-SR & HYPIR & TVT & GDPO-SR 
        & IDaS-SR \\
        \midrule
        {NFE $\downarrow$} 
        & 50 & 20 & 15 & 20 
        & 1 & 1 & 1 & 1 & 1 & 1 & 1 
        & 1 \\
        
        \midrule
        \multicolumn{13}{l}{\textit{DRealSR Dataset~\cite{drealsr}}} \\
        \midrule
        LPIPS$\downarrow$   & 0.314 & 0.350 & \secondbest{0.286} & 0.371 & 0.297 & 0.354 & 0.296 & 0.297 & 0.329 & 0.290 & \best{0.268} & 0.329 \\
        DISTS$\downarrow$   & 0.230 & 0.238 & \best{0.205} & 0.245 & 0.216 & 0.246 & 0.217 & 0.213 & 0.231 & 0.221 & \secondbest{0.211} & 0.239 \\
        % NIQE$\downarrow$    & 6.486 & 6.238 & \secondbest{5.887} & 6.701 & 6.447 & 5.985 & 6.172 & 5.915 & 6.414 & 7.032 & 6.535 & \best{5.741} \\
        % PI$\downarrow$      & 5.279 & 4.597 & 5.319 & 4.928 & 5.434 & \best{4.258} & 4.916 & 4.660 & 5.093 & 5.604 & 5.233 & \secondbest{4.541} \\
        CLIPIQA$\uparrow$ & 0.690 & 0.634 & 0.483 & 0.660 & 0.696 & 0.713 & 0.697 & \secondbest{0.735} & 0.627 & 0.722 & 0.703 & \best{0.746} \\
        % CLIPIQA+$\uparrow$& 0.676 & 0.656 & 0.538 & 0.665 & 0.675 & 0.693 & 0.692 & 0.676 & 0.647 & 0.666 & \secondbest{0.693} & \best{0.719} \\
        HyperIQA$\uparrow$& 0.658 & 0.612 & 0.502 & 0.604 & 0.605 & 0.615 & 0.639 & 0.613 & 0.577 & \best{0.666} & 0.616 & \secondbest{0.659} \\
        MUSIQ$\uparrow$   & 64.75 & 66.07 & 56.27 & 64.60 & 64.69 & 66.00 & 66.11 & \secondbest{66.60} & 61.47 & 65.56 & 65.59 & \best{66.81} \\
        % MANIQA$\uparrow$  & 0.602 & \best{0.640} & 0.513 & 0.623 & 0.590 & \secondbest{0.633} & 0.616 & 0.587 & 0.603 & 0.578 & 0.617 & 0.615 \\
        TOPIQ$\uparrow$     & 0.653 & 0.575 & 0.484 & 0.648 & 0.590 & 0.611 & 0.633 & 0.618 & 0.569 & \secondbest{0.659} & 0.603 & \best{0.674} \\
        
        \midrule
        \multicolumn{13}{l}{\textit{RealSR Dataset~\cite{realsr}}} \\
        \midrule
        LPIPS$\downarrow$   & 0.300 & 0.288 & 0.286 & 0.316 & 0.292 & 0.287 & \secondbest{0.267} & 0.272 & 0.305 & \best{0.260} & 0.268 & 0.325 \\
        DISTS$\downarrow$   & 0.222 & 0.211 & 0.208 & 0.223 & 0.213 & 0.212 & \secondbest{0.204} & 0.209 & 0.227 & 0.206 & \best{0.198} & 0.236 \\
        % NIQE$\downarrow$    & 5.391 & 5.378 & \best{4.743} & 5.877 & 5.649 & 5.628 & 5.504 & \secondbest{5.146} & 5.495 & 5.918 & 5.524 & 5.227 \\
        % PI$\downarrow$      & 4.074 & \secondbest{3.767} & 3.841 & 4.352 & 4.345 & 5.182 & 4.166 & \best{3.751} & 4.280 & 4.502 & 4.190 & 3.858 \\
        CLIPIQA$\uparrow$ & 0.667 & 0.615 & 0.529 & 0.628 & 0.669 & 0.679 & 0.670 & \secondbest{0.693} & 0.639 & 0.689 & 0.677 & \best{0.734} \\
        % CLIPIQA+$\uparrow$& 0.688 & 0.654 & 0.571 & 0.662 & 0.691 & 0.625 & 0.696 & 0.678 & 0.682 & 0.683 & \secondbest{0.703} & \best{0.745} \\
        HyperIQA$\uparrow$& 0.672 & 0.622 & 0.565 & 0.613 & 0.608 & 0.528 & 0.641 & 0.642 & 0.600 & \secondbest{0.676} & 0.617 & \best{0.680} \\
        MUSIQ$\uparrow$   & 69.70 & 68.77 & 65.06 & 67.96 & 69.08 & 68.54 & 70.15 & \best{70.72} & 66.25 & 69.89 & 69.07 & \secondbest{70.49} \\
        % MANIQA$\uparrow$  & 0.644 & \best{0.672} & 0.559 & 0.660 & 0.633 & \secondbest{0.662} & 0.655 & 0.634 & 0.652 & 0.623 & 0.617 & 0.652 \\
        TOPIQ$\uparrow$     & \secondbest{0.686} & 0.615 & 0.571 & 0.601 & 0.597 & 0.590 & 0.637 & 0.658 & 0.598 & 0.683 & 0.608 & \best{0.711} \\
        
        \midrule
        \multicolumn{13}{l}{\textit{RealPhoto60 Dataset~\cite{supir}}} \\
        \midrule
        % NIQE$\downarrow$    & 3.539 & 4.121 & 3.932 & \secondbest{3.331} & 3.755 & 4.647 & 3.564 & \best{3.155} & 3.436 & 4.644 & 3.630 & 3.813 \\
        % PI$\downarrow$      & 3.363 & 3.438 & 3.677 & 3.157 & 3.560 & 4.128 & 3.263 & \best{3.112} & \secondbest{3.124} & 4.118 & 3.378 & 3.373 \\
        CLIPIQA$\uparrow$ & 0.741 & 0.693 & 0.575 & 0.756 & 0.710 & 0.691 & 0.732 & \secondbest{0.758} & 0.730 & 0.755 & 0.739 & \best{0.796} \\
        % CLIPIQA+$\uparrow$& 0.726 & 0.687 & 0.602 & 0.705 & 0.718 & 0.673 & 0.726 & 0.730 & \secondbest{0.750} & 0.691 & 0.735 & \best{0.769} \\
        HyperIQA$\uparrow$& 0.648 & 0.583 & 0.557 & 0.639 & 0.619 & 0.585 & 0.642 & 0.635 & 0.636 & \secondbest{0.653} & 0.625 & \best{0.681} \\
        MUSIQ$\uparrow$   & 71.74 & 71.52 & 64.28 & 72.79 & 70.56 & 64.73 & 72.23 & \best{73.86} & 73.03 & 70.47 & 72.03 & \secondbest{73.65} \\
        TOPIQ$\uparrow$     & \secondbest{0.682} & 0.575 & 0.555 & 0.648 & 0.635 & 0.584 & 0.660 & 0.673 & 0.667 & 0.672 & 0.635 & \best{0.725} \\
        MANIQA$\uparrow$  & 0.611 & 0.403 & 0.528 & 0.621 & 0.590 & 0.563 & 0.614 & 0.600 & 0.592 & 0.594 & \best{0.629} & \secondbest{0.626} \\
        TReS$\uparrow$    & \secondbest{83.28} & 73.86 & 70.56 & 79.89 & 77.70 & 69.93 & 81.36 & 81.30 & 78.31 & 81.38 & 80.57 & \best{85.29} \\
        \bottomrule
    \end{tabular}
    \caption{Quantitative comparison on DRealSR, RealSR, and in-the-wild RealPhoto60 datasets. NFE denotes the number of diffusion sampling steps.
    }    \label{tab:quantitative_results}
\end{table*}

\vspace{0.1in} \noindent\textbf{Datasets and Metrics.}
Following previous works~\cite{seesr, osediff}, we trained on LSDIR~\cite{lsdir}, the first 10K face images from FFHQ~\cite{ffhq}, Flickr2K~\cite{flicker2k}, and DIV2K~\cite{div2k} using the degradation
pipeline of Real-ESRGAN~\cite{realesrgan} to synthesize LR-HR training pairs. 
For evaluation, we test on two paired real-world datasets- RealSR~\cite{realsr} and DRealSR~\cite{drealsr}, and an unpaired real-world dataset, RealPhoto60~\cite{supir}. 
We evaluate with LPIPS~\cite{lpips}, DISTS~\cite{dists}, CLIPIQA~\cite{clipiqa}, HyperIQA~\cite{hyperiqa}, MUSIQ~\cite{musiq}, TOPIQ~\cite{topiq}, MANIQA~\cite{maniqa}, and TReS~\cite{tres}.

\vspace{0.1in} \noindent\textbf{Implementation Details.}
In our experiments, we employ SD 2.1-base as the generative prior. The framework is optimized using AdamW. Specifically, the proposed MINE module utilizes a DiT-S/2 architecture~\cite{dit} and is trained with a learning rate of $1 \times 10^{-4}$, while the remaining network components are updated using a learning rate of $5 \times 10^{-5}$. To maintain computational efficiency, the LoRA rank is uniformly set to $r=8$ across the VAE encoder, the diffusion U-Net, and the regularizer. For robust semantic conditioning, we integrate the DAPE module~\cite{seesr} to facilitate efficient text prompt extraction. The objective function is balanced using the empirical weights $\lambda_{\text{LPIPS}}=2$, $\lambda_{\text{VSD}}=1$, and $\lambda_{\text{adv}}=0.1$. The complete model is trained with a total batch size of 2 for about 40 hours on two NVIDIA RTX 4090 GPUs. During inference, the generative steering parameter is set to $s=0.6$ by default to prioritize perceptual realism.
Throughout all tables, the best and second-best results are highlighted in \best{bolded} and \secondbest{underlined}, respectively.

\begin{figure*}[t!]
    \centering
    \includegraphics[width=\linewidth]{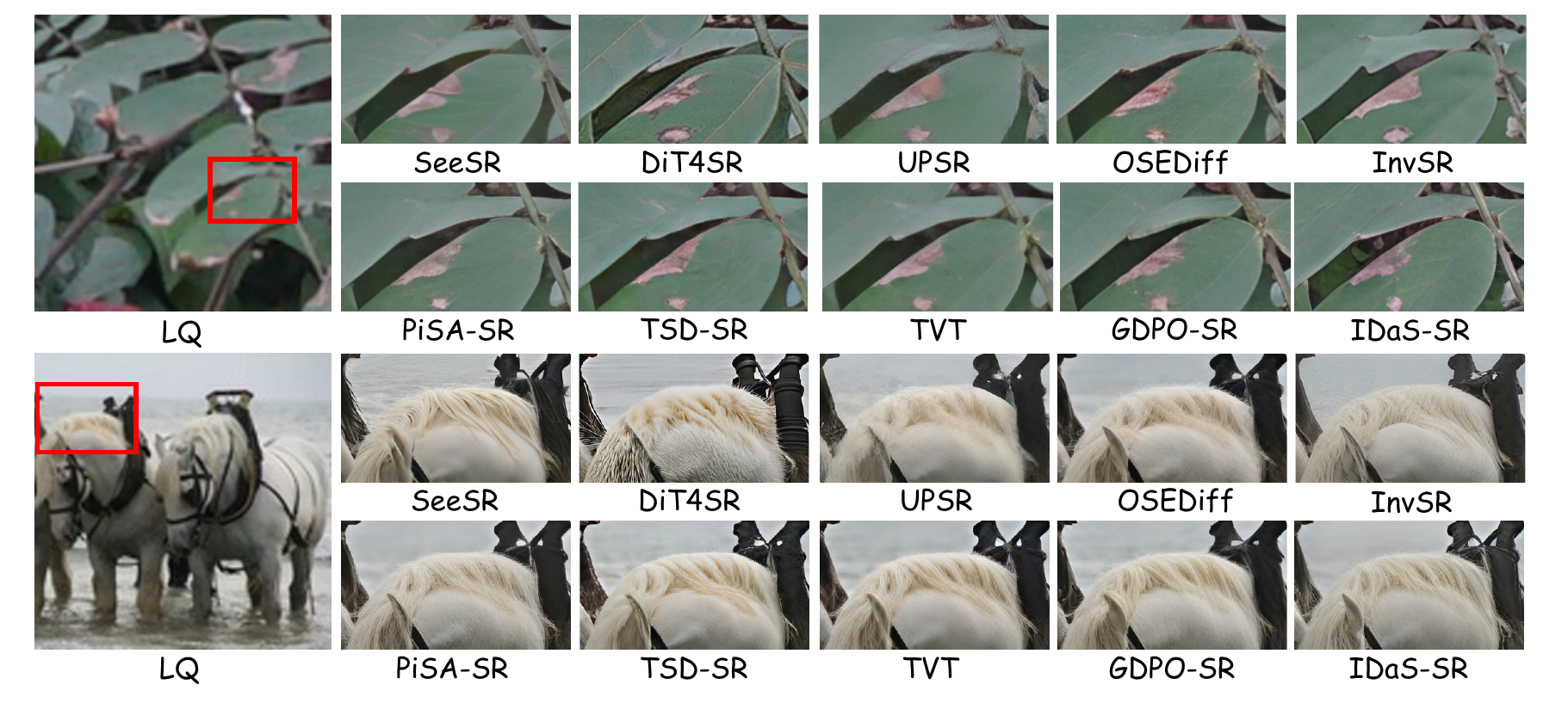}
    \caption{Visual comparison on real-world datasets~\cite{drealsr, supir}. Please zoom in for a better view.} 
    \label{fig:results}
\end{figure*}

\subsection{Comparison with State-of-the-Arts}

We evaluate IDaS-SR against eleven SOTA models~\cite{seesr, faithdiff, dit4sr, upsr, osediff, invsr, pisasr, hypir, tsdsr, tvt, gdposr} across three real-world datasets.

\noindent\textbf{Performance Comparison.}
As shown in Tab.~\ref{tab:quantitative_results}, IDaS-SR leads on most perceptual metrics with comparable fidelity performance across DRealSR and RealSR in a single step. 
On the in-the-wild RealPhoto60, it leads on most no-reference metrics, indicating generalization to real degradations unseen in training. The qualitative results illustrated in Fig.~\ref{fig:results} demonstrate that IDaS-SR restores sharper structure and more realistic texture than prior models.
% Complete comparisons are provided in the supplementary material.

\subsection{Ablation Study and Discussion}
In this section, we systematically ablate our core architecture, distillation objectives, and the generative control scalar. 

\noindent\textbf{Effectiveness of Core Components.}
Tab.~\ref{tab:ablation_modules} ablates IDaS-SR. Manifold Anchoring couples two roles: \emph{positioning} declares where the latent sits, through the anchoring timestep $\hat{t}$ and the degradation condition $\hat{c}_{\mathrm{deg}}$, while \emph{inversion} aligns the state with that position through the noise $\hat{\varepsilon}_{\mathrm{inv}}$. CHARIOT then traverses from this anchor.
Dropping $\hat{t}$ (a) defaults the anchor to $t{=}999$: fidelity survives because $\hat{\varepsilon}_{\mathrm{inv}}$ still encodes the input, but a structured state is treated as pure Gaussian noise and perceptual quality collapses. Dropping $\hat{c}_{\mathrm{deg}}$ (b) preserves fidelity yet lowers every perceptual metric, and dropping both (c) sends the latent off the trajectory, collapsing fidelity alongside realism as neither single removal does.
Inversion is the complementary requirement: without $\hat{\varepsilon}_{\mathrm{inv}}$ (d), the state no longer matches the position it declares, and the variant is dominated by the $s{=}0$ anchor on all metrics.
CHARIOT sits above the anchor rather than inside it. Removing it (e) leaves anchoring intact and lands almost exactly on the $s{=}0$ anchor, so the traversal adds the realism setting at no cost to the fidelity endpoint. Removing the latent GAN (f) keeps the traversal but weakens where it lands, lowering realism without recovering the fidelity the $s{=}0$ anchor already holds.
The full model thus attains the best perceptual scores across all metrics, while remaining adjustable down to $s{=}0$ when higher fidelity is preferred.

\begin{table}[t]
    \centering
    \small
    \setlength{\tabcolsep}{1.8mm}
    \begin{tabular}{c l ccc}
        \toprule
        \multicolumn{2}{l}{\textbf{Variant}} & LPIPS$\downarrow$ & CLIPIQA$\uparrow$ & HyperIQA$\uparrow$ \\
        \midrule
        (a) & w/o $\hat{t}$                        & 0.295 & 0.391 & 0.411 \\
        (b) & w/o $\hat{c}_{\mathrm{deg}}$          & \secondbest{0.293} & 0.692 & 0.603 \\
        (c) & w/o Positioning                       & 0.454 & 0.468 & 0.416 \\
        (d) & w/o $\hat{\varepsilon}_{\mathrm{inv}}$   & 0.305 & 0.679 & 0.589 \\
        (e) & w/o CHARIOT                           & \best{0.285} & 0.686 & 0.605 \\
        (f) & w/o l-GAN                             & 0.311 & \secondbest{0.725} & \secondbest{0.636} \\
        \midrule
        \multicolumn{2}{l}{IDaS-SR ($s{=}0$)}      & \best{0.285} & 0.689 & 0.610 \\
        \multicolumn{2}{l}{IDaS-SR (Ours)}         & 0.329 & \best{0.746} & \best{0.659} \\
        \bottomrule
\end{tabular}
    \caption{Ablation of the core components of IDaS-SR on DRealSR. Manifold anchoring pairs \emph{positioning} ($\hat{t}$, $\hat{c}_{\mathrm{deg}}$) with \emph{inversion} ($\hat{\varepsilon}_{\mathrm{inv}}$); (c) removes both positioning terms, while (e) and (f) remove the traversal of CHARIOT and the latent GAN (l-GAN), respectively. The generative steering parameter is set to $s=0.6$ by default.}
    \label{tab:ablation_modules}
\end{table}

\noindent\textbf{Positioning Requires Both Terms.}
Tab.~\ref{tab:positioning} ablates the two components of positioning, which act on the prior through distinct interfaces: the anchoring timestep $\hat{t}$ parameterizes the noise schedule and governs how much of the pretrained model is invoked, whereas the degradation condition $\hat{c}_{\mathrm{deg}}$ modulates the conditioning and specifies what is reconstructed at the resulting position.
Learning $\hat{t}$ alone provides a slight improvement over a fixed anchor, as a schedule coordinate cannot fully specify the position without a description of the observation.
Introducing $\hat{c}_{\mathrm{deg}}$ then improves every perceptual metric, and enabling both reaches the best perceptual quality at a modest cost in LPIPS.
Positioning is therefore a joint specification, not two separate refinements, which is why both coordinates are estimated jointly rather than tuned independently.

% \begin{table}[t]
%     \centering
%     \small
%     \setlength{\tabcolsep}{1.2mm}
%     \begin{tabular}{cc ccccc}
%         \toprule
%         $\hat{t}$ & $\hat{c}_{\mathrm{deg}}$ & LPIPS$\downarrow$ & DISTS$\downarrow$ & CLIP.$\uparrow$ & Hyper.$\uparrow$ & MUSIQ$\uparrow$ \\
%         \midrule
%         % Fixed    &            & \secondbest{0.301} & \secondbest{0.222} & 0.693 & 0.616 & 64.37 \\ %t=120
%         Fixed   &            & \secondbest{0.297} & \best{0.217} & 0.657 & 0.593 & 63.56 \\ %t=114
%         Learned &            & \best{0.293} & \secondbest{0.218} & 0.692 & 0.603 & 63.10 \\
%         % Fixed    & \checkmark & 0.316 & 0.230 & \secondbest{0.719} & \secondbest{0.632} & \secondbest{65.33} \\ %t=120
%         Fixed   & \checkmark & 0.308 & 0.226 & \secondbest{0.712} & \secondbest{0.632} & \secondbest{65.40} \\ % t=114
%         Learned & \checkmark & 0.329 & 0.239 & \best{0.746} & \best{0.659} & \best{66.81} \\
%         \bottomrule
%     \end{tabular}
%     \caption{Ablation of the positioning coordinates on DRealSR~\cite{drealsr}; all rows use the realism setting. The fixed anchor $t{=}114$ approximates the mean of the learned anchors ($\mu_{\hat{t}}{\approx}114$).}
%     \label{tab:positioning}
% \end{table}

\begin{table}[t]
    \centering
    \small
    \setlength{\tabcolsep}{1.8mm}
    \begin{tabular}{cc ccc}
        \toprule
        $\hat{t}$ & $\hat{c}_{\mathrm{deg}}$ & LPIPS$\downarrow$ & CLIPIQA$\uparrow$ & HyperIQA$\uparrow$ \\
        \midrule
        % Fixed    &            & \secondbest{0.301} & 0.693 & 0.616 \\ %t=120
        Fixed   &            & \secondbest{0.297} & 0.657 & 0.593 \\ %t=114
        Learned &            & \best{0.293} & 0.692 & 0.603 \\
        % Fixed    & \checkmark & 0.316 & \secondbest{0.719} & \secondbest{0.632} \\ %t=120
        Fixed   & \checkmark & 0.308 & \secondbest{0.712} & \secondbest{0.632} \\ % t=114
        Learned & \checkmark & 0.329 & \best{0.746} & \best{0.659} \\
        \bottomrule
    \end{tabular}
    \caption{Ablation of the positioning coordinates on DRealSR~\cite{drealsr}; all rows use the realism setting. The fixed anchor $t{=}114$ approximates the mean of the learned anchors ($\mu_{\hat{t}}{\approx}114$).}
    \label{tab:positioning}
\end{table}

\begin{figure}[t]
     \centering
     \includegraphics[width=\linewidth]{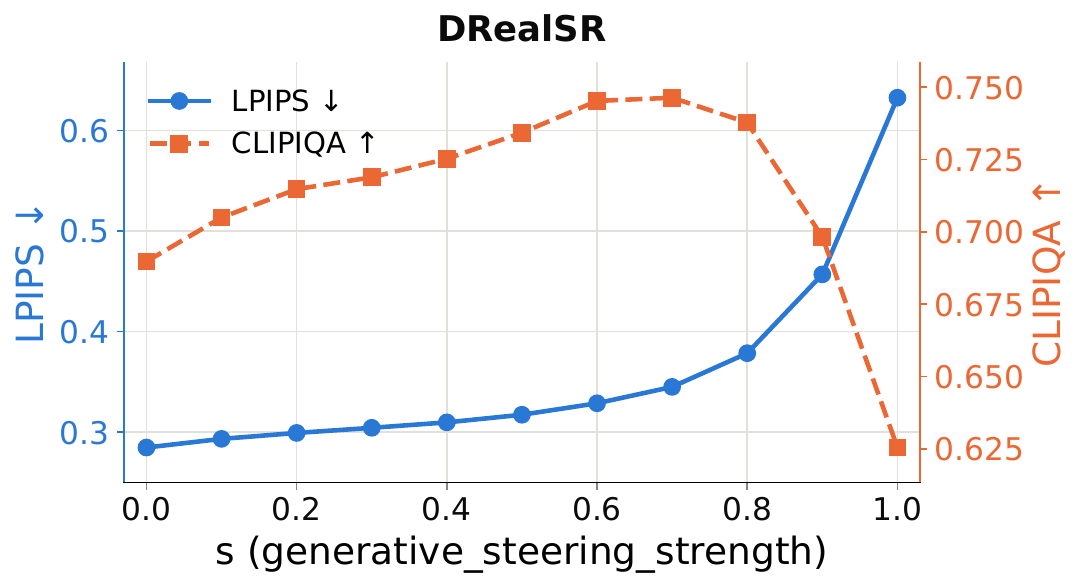}
     \caption{Fidelity-realism behavior along $s$ (LPIPS, CLIPIQA). Realism peaks near $s{=}0.6$; beyond it, the diluted $\hat{\varepsilon}_{\text{inv}}$ can no longer anchor the state to the input, so both metrics degrade as generation drifts from the observation.}
     \label{fig:steering_sweep}
\end{figure}

\noindent\textbf{Effectiveness of Generative Steering ($s$).}
Fig.~\ref{fig:steering_sweep} sweeps the control scalar on DRealSR. At the deterministic anchor ($s{=}0$), fidelity is highest but the restricted prior yields the lower perceptual score. 
Increasing $s$ activates the prior and lifts perceptual quality to a well-defined optimum near $s{=}0.6$, while LPIPS degrades monotonically; beyond the optimum, unconstrained stochasticity overrides the anchor and both fidelity and perceptual quality deteriorate. 
This single-peaked response yields a stable operating range $s\in[0,0.6]$, over which one set of weights navigates the fidelity-realism boundary from restorer toward synthesizer.

\section{Conclusion}
In this work, we present IDaS-SR, a one-step diffusion framework that bridges deterministic restoration and controllable generation for Real-ISR. 
The key idea is that restoration and generation need not be opposing objectives: by anchoring the degraded input to a well-defined point on the pretrained trajectory and then steering from that anchor, IDaS-SR keeps the latent faithful to the input even as it moves toward generation. 
By maintaining this consistency throughout the traversal, it steers toward realistic generation while avoiding the instability typical of prior one-step approaches. 
Extensive experiments demonstrate that IDaS-SR achieves leading perceptual realism among one-step methods, offering continuous fidelity-realism control from a single scalar.

{
    \small
    \bibliographystyle{ieeenat_fullname}
    \bibliography{main}
}

\clearpage
\appendix
\renewcommand{\theequation}{A\arabic{equation}}
\renewcommand{\thetable}{A\arabic{table}}
\renewcommand{\thefigure}{A\arabic{figure}}
\renewcommand{\thealgorithm}{A\arabic{algorithm}}
\setcounter{equation}{0}
\setcounter{table}{0}
\setcounter{figure}{0}
\setcounter{algorithm}{0}
\setcounter{section}{0}

\twocolumn[{%
  \centering
  {\Large\bf Supplementary Material}\\[1em]
}]

\section{Detailed Methodology and Implementation}
This section details the formal derivations and implementation settings of IDaS-SR, structured to mirror the end-to-end pipeline. Sec.~\ref{sec.explain_MINE}--\ref{sec.explain_positioning} elaborate the Manifold Inversion Noise Estimator (MINE): its feature extraction and its estimation of the two positioning coordinates. Sec.~\ref{sec.explain_CHARIOT} details CHARIOT, which governs the controlled generative traversal.

\subsection{Decoupling Trajectory via Manifold Inversion Noise Estimator}
\label{sec.explain_MINE}
MINE mitigates the ill-posed nature of manifold inversion by decoupling the estimation of local spatial noise from the global positioning context. As shown in Fig.~2 of the main paper, extraction is driven by a {DiT} backbone and {MLP} heads. Given a low-quality latent $z_L$ and learnable query tokens $\tau_{query}$:
\begin{equation}
    \bar{p},\, \hat{\epsilon}_{inv} = \text{DiT}(\tau_{query}, z_L),
\end{equation}
where the inversion noise $\hat{\epsilon}_{inv}$ projects the LQ latent onto the diffusion trajectory to address the distributional mismatch, and the pooled vector $\bar{p}$ serves as a global context representation. MINE routes $\bar{p}$ through two MLP heads to derive the anchoring timestep $\hat{t}$ and the degradation embedding $\hat{c}_{deg}$.

Architecturally, $z_L \in \mathbb{R}^{B \times C \times H \times W}$ is partitioned into non-overlapping patches and linearly projected to a spatial sequence $T_{spatial} \in \mathbb{R}^{B \times N \times 384}$. To extract holistic image properties—structural complexity and degradation severity—without corrupting local features, we prepend $K=4$ learnable query tokens $\tau_{query} \in \mathbb{R}^{B \times K \times 384}$. After the transformer, the output is decoupled: the $N$ spatial tokens are projected and unpatchified to reconstruct the grid, yielding $\hat{\epsilon}_{inv} \in \mathbb{R}^{B \times C \times H \times W}$; the query tokens are condensed into $\bar{p} \in \mathbb{R}^{B \times 384}$ by global average pooling. This separation also regularizes: by routing global semantics through a compressed pooled vector while confining high-frequency residuals to the dense spatial grid, the network is prevented from finding trivial solutions without auxiliary losses.

\subsection{Estimation of Positioning Coordinates}
\label{sec.explain_positioning}
The pooled context $\bar{p}$ is routed through two lightweight heads to derive the two positioning coordinates: the anchoring timestep $\hat{t}$ and the degradation embedding $\hat{c}_{deg}$.

\noindent\textbf{Anchoring Timestep ($\hat{t}$):}
$\hat{t}$ locates the restoration starting point on the trajectory. We formulate it as a continuous variable to permit end-to-end optimization across the discrete schedule. The context $\bar{p}$ is processed by $\text{MLP}_{t}$ and bounded by a sigmoid:
\begin{equation}
    \hat{t} = t_{min} + (t_{max} - t_{min}) \cdot \sigma(\text{MLP}_{t}(\bar{p})),
\end{equation}
where $t_{min}=50$ and $t_{max}=450$ bound a valid structural anchor. The clamp also regularizes: confining $\hat{t}$ to this sub-manifold eliminates timestep explosion or collapse under unconstrained implicit optimization.

For inversion we define a continuous operative timestep $\tilde{t}$, denoting either $\hat{t}$ or the rescheduled $t_{mix}$ (when CHARIOT is active). Since the pretrained prior relies on a discrete schedule, we map $\tilde{t}$ to schedule parameters via differentiable piecewise-linear interpolation:
\begin{equation}
    \alpha_{\tilde{t}} = \alpha_{\lfloor \tilde{t} \rfloor} + (\tilde{t} - \lfloor \tilde{t} \rfloor)(\alpha_{\lceil \tilde{t} \rceil} - \alpha_{\lfloor \tilde{t} \rfloor}),
    \label{eq.schedule_param}
\end{equation}
\begin{equation}
    \beta_{\tilde{t}} = \beta_{\lfloor \tilde{t} \rfloor} + (\tilde{t} - \lfloor \tilde{t} \rfloor)(\beta_{\lceil \tilde{t} \rceil} - \beta_{\lfloor \tilde{t} \rfloor}).
    \label{eq.schedule_param_beta}
\end{equation}

This keeps $z_{\tilde{t}} = \alpha_{\tilde{t}}z_L + \beta_{\tilde{t}}\hat{\epsilon}_{inv}$ differentiable, so gradients from the primary objectives flow through the U-Net and the interpolation weights back to MINE. Protected by these bottlenecks and bounds, this implicit supervision lets the network discover anchors stably without ground-truth timestep labels.

\noindent\textbf{Degradation Embedding $(\hat{c}_{deg})$:} To supply explicit degradation context, the pooled vector $\bar{p}$ is projected to the semantic dimension ($d_{ctx}=1024$) of the pretrained model via $\text{MLP}_{cond}$:
\begin{equation}
    \hat{c}_{deg} = \text{MLP}_{cond}(\bar{p}) \in \mathbb{R}^{B \times d_{ctx}}.
\end{equation}

The embedding is then broadcast across the token axis and added element-wise to the CLIP text embeddings $c_y$:
\begin{equation}
    \tilde{c}_y = c_y + \mathbf{1}_{S} \otimes \hat{c}_{deg} \in \mathbb{R}^{B \times S \times d_{ctx}},
\end{equation}
where $S=77$ is the text sequence length and $\mathbf{1}_{S}$ broadcasts across tokens. This additive bias shifts the prompt to modulate generative guidance for the degradation severity, carrying the LQ-to-HQ relation that a scalar cannot express.

\subsection{Continuous Rescheduling and Variance Normalization in CHARIOT}
\label{sec.explain_CHARIOT}
CHARIOT injects stochastic noise into the deterministic inverted latent while rescheduling the starting timestep, both governed by the steering scalar $s \in [0, 1)$. Sampled from $\mathcal{U}(0,1)$ in training and set deterministically at inference, $s$ first extends $\hat{t}$ to:
\begin{equation}
    t_{mix} = \hat{t} + s \cdot (T_{max} - \hat{t}), \quad T_{max}=1000.
\end{equation}

Concurrently, $s$ sets the noise mixture, transitioning from the structural anchor ($s{=}0$) toward the generative prior ($s{\to}1$):
\begin{equation}
    \epsilon_{raw} = (1 - s)\hat{\epsilon}_{inv} + s\,\epsilon, \quad \epsilon \sim \mathcal{N}(0, \mathbf{I}).
\end{equation}

Mixing two independent unit-variance fields shrinks the variance by $\sqrt{(1-s)^2+s^2}<1$. Rescaling strictly to unit variance, however, would prematurely snap the latent onto the standard Gaussian trajectory, over-amplifying the low-energy structure of $\hat{\epsilon}_{inv}$ (e.g., in flat regions). We therefore define $\text{Norm}(\cdot)$ from Eq.~(8) of the main paper to rescale toward an interpolated target energy rather than to $1$:
\begin{equation}
    \sigma_{target} = (1 - s)\,\text{std}(\hat{\epsilon}_{inv}) + s,
\end{equation}
which reaches unit variance only at $s{=}1$ and otherwise preserves a fraction of the inversion-noise energy. Here $\text{std}(\cdot)$ is computed per sample across $C \times H \times W$, retaining the spatial patterns of $\hat{\epsilon}_{inv}$ rather than imposing a rigid global scale. The mixture is then rescaled as:
\begin{equation}
    \hat{\epsilon}_{mix} = \text{Norm}(\epsilon_{raw}) = \epsilon_{raw} \cdot \frac{\sigma_{target}}{\text{std}(\epsilon_{raw}) + \delta},
\end{equation}
where $\delta = 10^{-8}$. 
Dividing by $\text{std}(\epsilon_{raw})$ removes the shrinkage introduced by mixing, while scaling to $\sigma_{target}$ rather than to $1$ preserves the low-energy structure of $\hat{\epsilon}_{inv}$ at small $s$. 
Adjusting the noise level therefore never washes out local fidelity. 
The steered latent follows:
\begin{equation}
    z_{t_{mix}} = \alpha_{t_{mix}}z_{L} + \beta_{t_{mix}}\hat{\epsilon}_{mix},
\end{equation}
with $\alpha_{t_{mix}}, \beta_{t_{mix}}$ from Eq.~\eqref{eq.schedule_param}. Because the noise mixture and $t_{mix}$ share the same $s$, the endpoints stay consistent: at $s{=}0$ the latent is the deterministic anchor, and at the generative limit the noise reaches unit variance matching $T_{max}$. For intermediate $s$, the injected noise is deliberately weaker than the nominal marginal at $t_{mix}$, keeping the traversal anchored to the input rather than snapping to the standard trajectory.

\subsection{Overall Algorithm of IDaS-SR}
To synthesize the detailed mechanisms of MINE and CHARIOT, we provide the complete training procedure of IDaS-SR in Alg.~\ref{alg:idas_sr}. The pseudo-code shows how a low-quality input is deterministically anchored, stochastically rescheduled, and supervised so that a single diffusion step yields a high-fidelity, perceptually realistic image at inference.

\begin{table*}[t]
    \centering
    \small
    \setlength{\tabcolsep}{1.2mm}
    \begin{tabular}{l cc ccccccccc}
        \toprule
        \multirow{2}{*}{Metric} & \multicolumn{2}{c}{Multi-Step Diffusion} & \multicolumn{9}{c}{One-Step Diffusion} \\
        \cmidrule(lr){2-3} \cmidrule(lr){4-12}
        & SeeSR & UPSR & SinSR & OSEDiff & InvSR & PiSA-SR & TSD-SR & HYPIR & TVT & CODSR & \textbf{IDaS-SR} \\
        \midrule
        Steps                 & 200     & 5      & 1      & 1       & 1       & 1       & 1      & 1      & 1      & 1      & 1 \\
        Time (ms)$\downarrow$ & 7902.00 & 562.76 & 286.60 & 197.24  & 257.76  & 220.31  & 228.00 & 998.07 & 697.00 & 242.00 & 263.46 \\
        Total Params (M)      & 2510.76 & 121.95 & 173.92 & 1289.95 & 1323.79 & 1289.95 & 2207.30& 1549.43& 1256.00& 1293.00& 1323.81 \\
        Trainable Params (M)  & 749.88  & 119.42 & 173.92 & 340.39  & 33.80   & 8.11    & 95.20  & 259.47 & 34.60  & 35.80  & 373.85 \\
        \bottomrule
    \end{tabular}
    \caption{Complexity and inference speed comparison. All metrics are evaluated on a single NVIDIA RTX 2080 Ti for $4\times$ SR ($512 \times 512$ output resolution).}
    \label{tab:efficiency}
\end{table*}

\begin{algorithm}[tb]
\caption{Training Scheme of IDaS-SR}
\label{alg:idas_sr}
\textbf{Input}: Dataset $\mathcal{S}$; frozen SD $(\mathcal{E}, \mathcal{D}, \epsilon_{\theta_{pre}})$; prompt extractor $Y$; iterations $N$; generative limit $T_{max}$\\
\textbf{Output}: Encoder $\mathcal{E}_\theta$, MINE $f_\phi$, UNet $\epsilon_\theta$
\begin{algorithmic}[1]
\STATE \textbf{Init:} LoRA weights $\theta$ (UNet), $\theta_{\mathcal{E}}$ (encoder), $\theta_{ft}$ (VSD regularizer); MINE $f_\phi$; discriminator $D_\psi$
\FOR{$i = 1$ \TO $N$}
    \STATE Sample $(x_L, x_H) \sim \mathcal{S}$; \; $c_y \gets Y(x_L)$
    \STATE $z_L \gets \mathcal{E}_\theta(x_L)$; \; $z_H \gets \mathcal{E}_\theta(x_H)$
    \STATE \textit{// Manifold anchoring (MINE)}
    \STATE $(\hat{\epsilon}_{inv}, \bar{p}) \gets f_\phi(z_L)$; \; $\hat{t} \gets \text{MLP}_t(\bar{p})$; \; $\hat{c}_{deg} \gets \text{MLP}_{c}(\bar{p})$
    \STATE \textit{// Controlled traversal (CHARIOT)}
    \STATE Sample $s \sim \mathcal{U}(0, 1)$, \; $\epsilon_{r} \sim \mathcal{N}(0, I)$
    \STATE $t_{mix} \gets \hat{t} + s\,(T_{max} - \hat{t})$; \; interpolate $\alpha_{t_{mix}}, \beta_{t_{mix}}$
    \STATE $\hat{\epsilon}_{mix} \gets \text{Norm}\big((1-s)\hat{\epsilon}_{inv} + s\,\epsilon_{r}\big)$
    \STATE \textit{// One-step denoising}
    \STATE $z_{t_{mix}} \gets \alpha_{t_{mix}} z_L + \beta_{t_{mix}} \hat{\epsilon}_{mix}$
    \STATE $\hat{z}_H \gets \big(z_{t_{mix}} - \beta_{t_{mix}}\, \epsilon_\theta(z_{t_{mix}}; t_{mix}, c_y{+}\hat{c}_{deg})\big) / \alpha_{t_{mix}}$
    \STATE $\hat{x}_H \gets \mathcal{D}(\hat{z}_H)$
    \STATE \textit{// Reconstruction loss}
    \STATE $\mathcal{L}_{data} \gets \mathcal{L}_{MSE}(\hat{x}_H, x_H) + \lambda_{lpips}\,\mathcal{L}_{LPIPS}(\hat{x}_H, x_H)$
    \STATE \textit{// Latent adversarial loss (noisy discrimination)}
    \STATE Sample $t_d \sim \mathcal{U}(1, 400)$, \; $\epsilon_d \sim \mathcal{N}(0, I)$
    \STATE $\tilde{z}_H \gets \alpha_{t_d}\,\text{sg}[z_H] + \beta_{t_d}\epsilon_d$; \; $\tilde{z}_{\hat{H}} \gets \alpha_{t_d}\hat{z}_H + \beta_{t_d}\epsilon_d$
    \STATE $\mathcal{L}_D \gets \mathbb{E}[\max(0, 1{-}D_\psi(\tilde{z}_H, t_d))] + \mathbb{E}[\max(0, 1{+}D_\psi(\text{sg}[\tilde{z}_{\hat{H}}], t_d))]$
    \STATE $\mathcal{L}_{adv} \gets -\,\mathbb{E}[D_\psi(\tilde{z}_{\hat{H}}, t_d)]$
    \STATE \textit{// Score distillation loss}
    \STATE Sample $t \sim \mathcal{U}(1, T_{max})$, \; $\epsilon \sim \mathcal{N}(0, I)$; \; $z_t \gets \alpha_t \hat{z}_H + \beta_t \epsilon$
    \STATE $\mathcal{L}_{VSD} \gets \mathbb{E}\big[\|\epsilon_{\theta_{pre}}(z_t; t, c_y) - \epsilon_{\theta_{ft}}(z_t; t, c_y)\|^2\big]$
    \STATE $\mathcal{L}_{diff} \gets \mathcal{L}_{MSE}\big(\epsilon_{\theta_{ft}}(\text{sg}[z_t]; t, c_y),\, \epsilon\big)$
    \STATE \textit{// Parameter updates (SD backbone frozen)}
    \STATE $\mathcal{L}_{total} \gets \mathcal{L}_{data} + \lambda_{VSD}\mathcal{L}_{VSD} + \lambda_{adv}\mathcal{L}_{adv}$
    \STATE $\{\theta, \theta_{\mathcal{E}}, \phi\} \xleftarrow{\nabla} \mathcal{L}_{total}$; \; $\psi \xleftarrow{\nabla} \mathcal{L}_D$; \; $\theta_{ft} \xleftarrow{\nabla} \mathcal{L}_{diff}$
\ENDFOR
\end{algorithmic}
\end{algorithm}

\section{Additional Experiments and Analysis}
\label{sec:extended_ablation}

\subsection{Extended Quantitative Comparison}
\label{sec:extended_results}
We present a detailed comparison against prior methods~\cite{seesr, faithdiff, upsr, dit4sr, sinsr, osediff, invsr, pisasr, tsdsr, hypir, tvt, gdposr, codsr} on the real-world RealSR~\cite{realsr} and DRealSR~\cite{drealsr} benchmarks and on DIV2K-Val~\cite{div2k}, where LQ inputs are synthesized with the Real-ESRGAN~\cite{realesrgan} degradation pipeline.
Tab.~\ref{tab:extended_full} reports the complete metric set at the realism default ($s{=}0.6$), where IDaS-SR leads on perceptual quality across all three benchmarks. 
Tab.~\ref{tab:fidelity} additionally compares IDaS-SR at its fidelity anchor ($s{=}0$) against recent one-step methods on full-reference metrics, where it ranks among the top methods, showing that the same model also achieves strong fidelity when steered toward $s{=}0$.
Figs.~\ref{fig:visual-1} and \ref{fig:visual-2} provide qualitative comparisons across datasets. By bridging restoration and generation within a single diffusion step, IDaS-SR recovers faithful, photo-realistic detail, balancing the structural and generative manifolds to exploit the pretrained diffusion prior.

\begin{table}[t]
    \centering
    \small
    \setlength{\tabcolsep}{2mm}
    \begin{tabular}{c ccc}
        \toprule
        \textbf{Control} ($s$) & LPIPS$\downarrow$ & CLIPIQA$\uparrow$ & HyperIQA$\uparrow$ \\
        \midrule
        0.0 & \best{0.285} & 0.690 & 0.608 \\
        0.1 & \secondbest{0.293} & 0.705 & 0.631 \\
        0.2 & 0.299 & 0.715 & 0.642 \\
        0.3 & 0.304 & 0.719 & 0.650 \\
        0.4 & 0.310 & 0.725 & \secondbest{0.654} \\
        0.5 & 0.317 & 0.734 & \best{0.657} \\
        \textbf{0.6} & 0.329 & \secondbest{0.745} & \best{0.657} \\
        0.7 & 0.345 & \best{0.746} & \secondbest{0.654} \\
        0.8 & 0.379 & 0.738 & 0.644 \\
        0.9 & 0.457 & 0.698 & 0.639 \\
        1.0 & 0.633 & 0.626 & 0.619 \\
        \bottomrule
    \end{tabular}
    \caption{Effect of the generative steering parameter $s$ on DRealSR~\cite{drealsr}. The default setting of our method is $s{=}0.6$. Best and second are \best{bolded} and \secondbest{underlined}, respectively.}
    \label{tab:ablation_steering}
\end{table}

\begin{figure}[t]
    \centering
    \includegraphics[width=\linewidth]{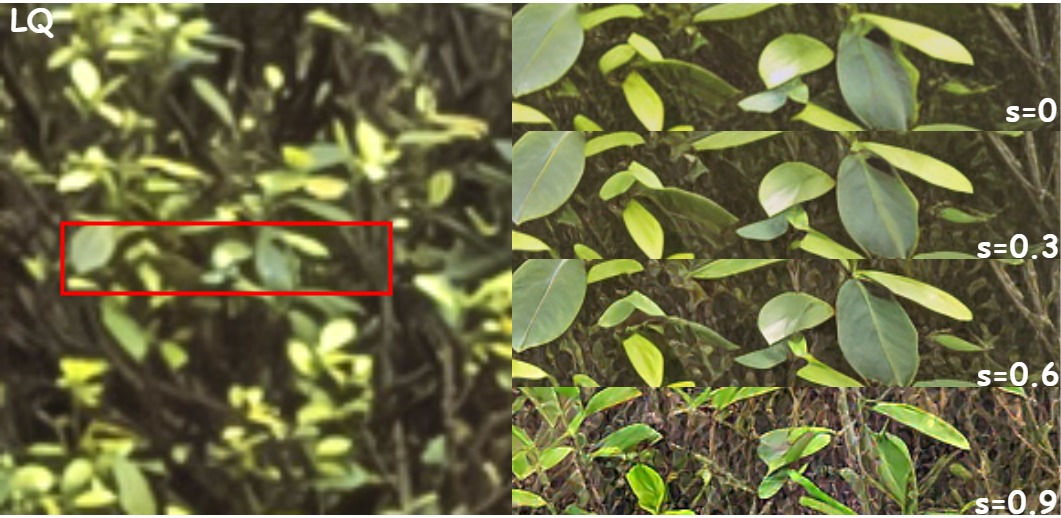}
    \caption{Qualitative trajectory of IDaS-SR under CHARIOT steering, from the structural anchor ($s{=}0$) to the generative regime. The generative steering parameter is set to $s {=} 0.6$ by default.}
    \label{fig:ablation_chariot}
\end{figure}

\begin{table}[t]
    \centering
    \small
    \setlength{\tabcolsep}{1.2mm}
    \begin{tabular}{l c cc cc}
        \toprule
        \textbf{Method} & \textbf{NFE}$\downarrow$ & \textbf{PSNR}$\uparrow$ & \textbf{SSIM}$\uparrow$ & \textbf{LPIPS}$\downarrow$ & \textbf{DISTS}$\downarrow$ \\
        \midrule
        \multicolumn{6}{l}{\textit{DIV2K-Val Dataset~\cite{div2k}}} \\
        \cmidrule(lr){1-6}
        OSEDiff         & 1  & 23.72 & \thirdbest{0.611} & 0.294 & 0.198 \\
        InvSR           & 1  & 22.90 & 0.591 & 0.319 & 0.206 \\
        % PiSA-SR         & 1  & 23.87 & 0.606 & 0.282 & 0.193 \\
        TSD-SR          & 1  & 23.02 & 0.581 & \best{0.267} & \best{0.182} \\
        HYPIR           & 1  & 22.21 & 0.576 & 0.308 & 0.195 \\
        GDPO-SR         & 1  & \best{23.92} & \secondbest{0.612} & 0.290 & 0.196 \\
        CODSR           & 1  & \secondbest{23.79} & 0.607 & \thirdbest{0.288} & \thirdbest{0.193} \\
        \cmidrule(lr){1-6}
        IDaS-SR ($s{=}0$) & 1 & \thirdbest{23.73} & \best{0.617} & \secondbest{0.285} & \secondbest{0.192} \\
        
        \midrule
        \multicolumn{6}{l}{\textit{DRealSR Dataset~\cite{drealsr}}} \\
        \cmidrule(lr){1-6}
        OSEDiff         & 1  & 27.92 & \best{0.784} & 0.297 & \thirdbest{0.216} \\
        InvSR           & 1  & 25.67 & 0.713 & 0.354 & 0.246 \\
        % PiSA-SR         & 1  & 28.32 & 0.780 & 0.296 & 0.217 \\
        TSD-SR          & 1  & 27.77 & 0.756 & 0.297 & \secondbest{0.213} \\
        HYPIR           & 1  & 26.07 & 0.728 & 0.329 & 0.231 \\
        GDPO-SR         & 1  & \best{28.21} & 0.734 & \best{0.268} & \best{0.211} \\
        CODSR           & 1  & \secondbest{28.19} & \thirdbest{0.776} & \thirdbest{0.292} & \best{0.211} \\
        \cmidrule(lr){1-6}
        IDaS-SR ($s{=}0$) & 1 & 27.49 & \secondbest{0.777} & \secondbest{0.285} & \secondbest{0.213} \\
        
        \midrule
        \multicolumn{6}{l}{\textit{RealSR Dataset~\cite{realsr}}} \\
        \cmidrule(lr){1-6}
        OSEDiff         & 1  & \thirdbest{25.15} & \best{0.734} & 0.292 & 0.213 \\
        InvSR           & 1  & 24.13 & 0.713 & 0.287 & 0.212 \\
        % PiSA-SR         & 1  & 25.50 & 0.742 & 0.267 & 0.204 \\
        TSD-SR          & 1  & 24.82 & 0.712 & \secondbest{0.272} & 0.209 \\
        HYPIR           & 1  & 22.88 & 0.686 & 0.305 & 0.227 \\
        GDPO-SR         & 1  & \best{25.50} & \best{0.734} & \best{0.268} & \best{0.198} \\
        CODSR           & 1  & \secondbest{25.37} & \secondbest{0.728} & \thirdbest{0.274} & \secondbest{0.200} \\
        \cmidrule(lr){1-6}
        IDaS-SR ($s{=}0$) & 1 & 24.71 & 0.725 & \thirdbest{0.274} & \thirdbest{0.204} \\
        \bottomrule
    \end{tabular}
    \caption{Fidelity comparison among one-step methods across DIV2K-Val~\cite{div2k}, DRealSR~\cite{drealsr}, and RealSR~\cite{realsr}. At its fidelity anchor ($s{=}0$), IDaS-SR achieves full-reference scores competitive with distortion-oriented methods. Best, second, and third are \best{bolded}, \secondbest{underlined}, and \thirdbest{italicized}, respectively.}
    \label{tab:fidelity}
\end{table}

\subsection{Complexity Comparison}
Tab.~\ref{tab:efficiency} compares inference cost. As a one-step model, IDaS-SR is roughly $30\times$ faster than the multi-step SeeSR and comparable to accelerated baselines like UPSR. Its runtime is competitive among one-step methods and well below the heavier variants (HYPIR, TVT). MINE adds trainable parameters relative to lightweight adapters, reflecting the cost of jointly estimating the anchor, degradation embedding, and inversion noise; this budget buys the manifold alignment behind IDaS-SR's leading perceptual quality while keeping single-step runtime on par with far simpler architectures.

\subsection{Analysis of CHARIOT Generative Steering}
The main paper analyzes the fidelity--realism trade-off of CHARIOT along $s$ in Fig.~4. Here we provide a finer quantitative sweep (Tab.~\ref{tab:ablation_steering}) and a qualitative visualization (Fig.~\ref{fig:ablation_chariot}) across the steering range.

As $s$ increases from the deterministic anchor toward the generative prior, the model transitions from strict structural fidelity to synthesized realism. At $s{=}0$, it preserves the input layout with the best full-reference scores but lacks high-frequency detail. Increasing $s$ activates the generative prior to synthesize realistic textures (e.g., leaf veins), with perceptual metrics peaking near $s{=}0.6$. Beyond this, at $s{=}0.9$, the trade-off tilts sharply toward realism: synthesized details remain perceptually plausible but diverge from the ground-truth structure, and full-reference fidelity degrades steeply. We therefore recommend $s \in [0, 0.6]$ for practical restoration.

Although inference stays within this range, we train across the full $s \in [0, 1]$ so the anchor is optimized at every operating point rather than at a single setting. This full-range exposure aligns the interpolation trajectory with the pretrained manifold, keeping low-$s$ inference on-distribution while the generative prior remains available at higher $s$.

\subsection{Analysis of MINE Positioning}
\label{sec:positioning_analysis}
Positioning comprises the anchoring timestep $\hat{t}$ and the degradation embedding $\hat{c}_{deg}$ (Sec.~4.3). We analyze what each encodes.

\begin{figure}[t]
    \centering
    \begin{subfigure}[b]{0.48\linewidth}
        \centering
        \includegraphics[width=\linewidth]{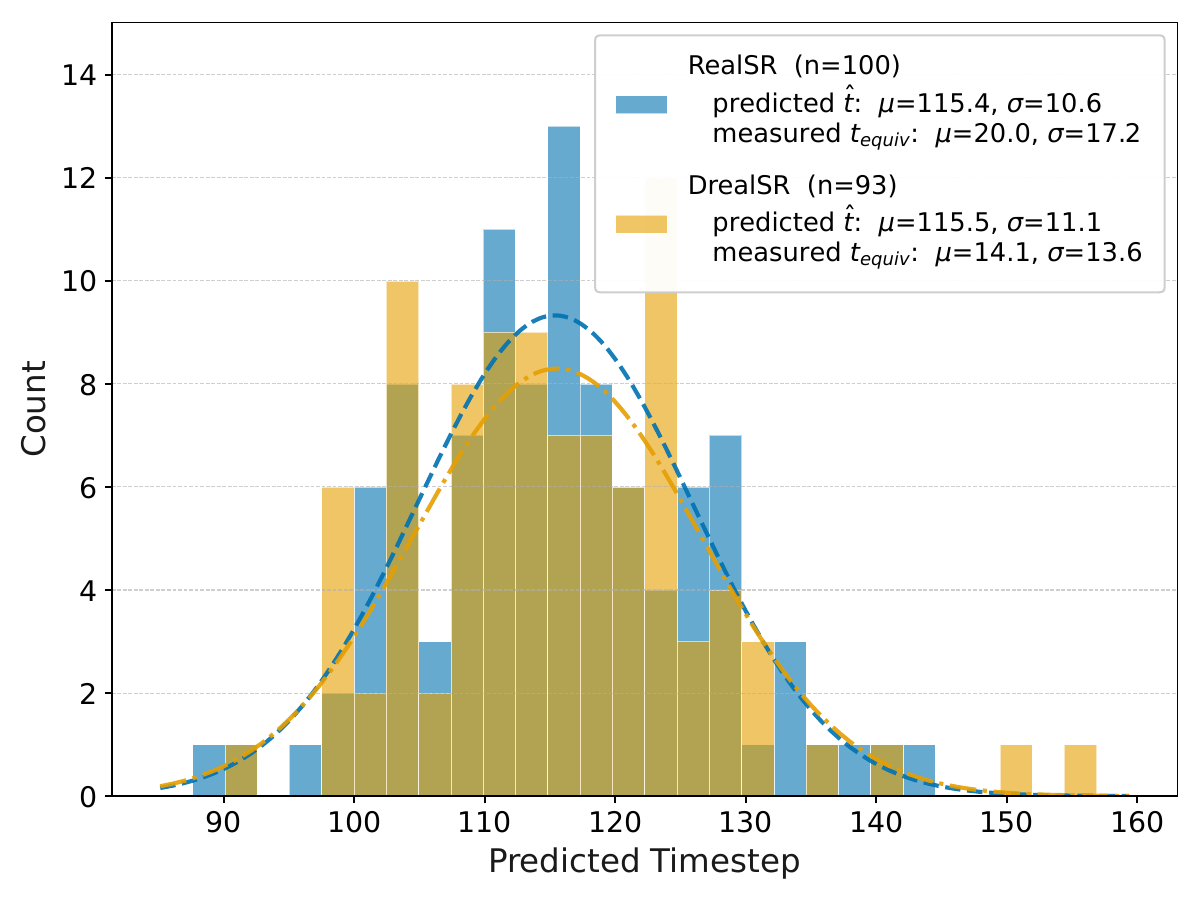}
        \caption{Anchoring timestep $\hat{t}$}
        \label{fig:timestep_histogram}
    \end{subfigure}
    \hfill
    \begin{subfigure}[b]{0.48\linewidth}
        \centering
        \includegraphics[width=\linewidth]{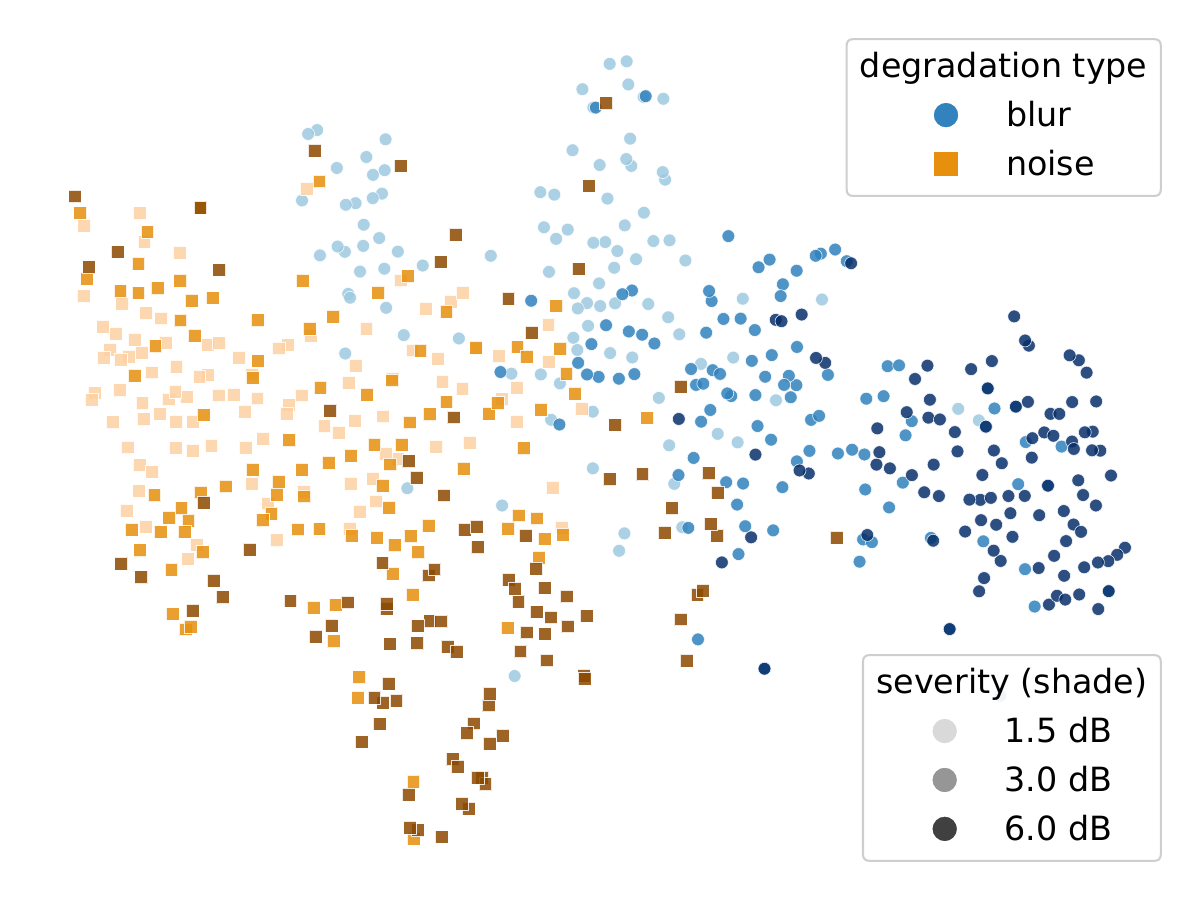}
        \caption{Degradation embedding $\hat{c}_{deg}$}
        \label{fig:tsne_cdeg}
    \end{subfigure}
    \caption{Analysis of MINE's positioning. (a) On RealSR~\cite{realsr} and DRealSR~\cite{drealsr}, the LQ latents are near-clean ($t_{equiv}\!\approx\!14$--$20$) while MINE anchors them deeper at a consistent $\hat{t}\!\approx\!115$ ($\bar{\alpha}\!\approx\!0.85$, $\sigma\!\approx\!11$)---far enough to engage the generative prior while preserving structure. (b) t-SNE of $\hat{c}_{deg}$ on a balanced grid of blur/noise at three severities (content removed by within-image centering): degradation type forms distinct clusters and severity a within-cluster gradient.}
    \label{fig:mine_analysis}
\end{figure}

\begin{figure*}[p]
    \centering
    \includegraphics[width=\linewidth]{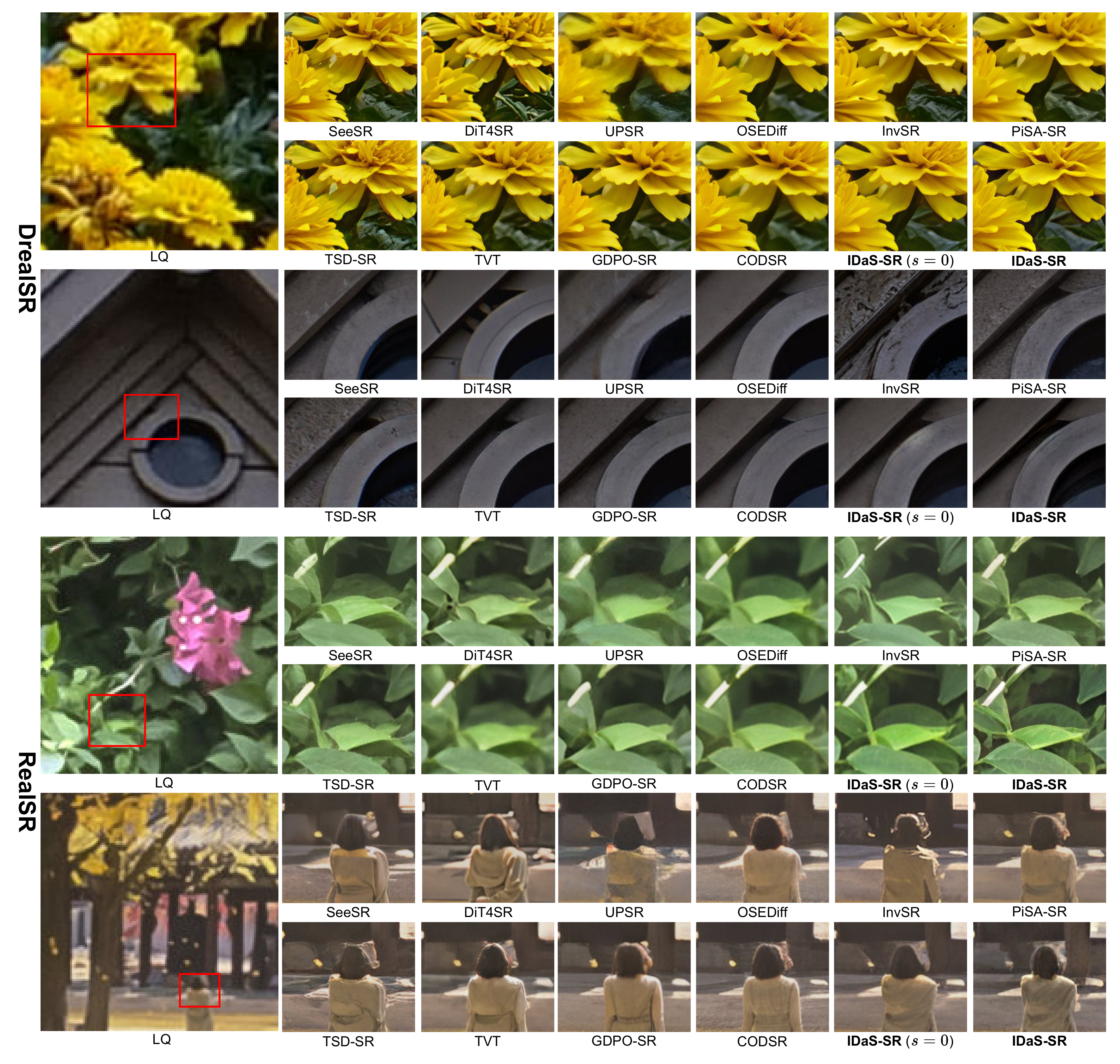}
    \caption{Qualitative comparisons on the RealSR~\cite{realsr} and DrealSR~\cite{drealsr} datasets.}
    \label{fig:visual-1}
\end{figure*}

\begin{figure*}[p]
    \centering
    \includegraphics[width=\linewidth]{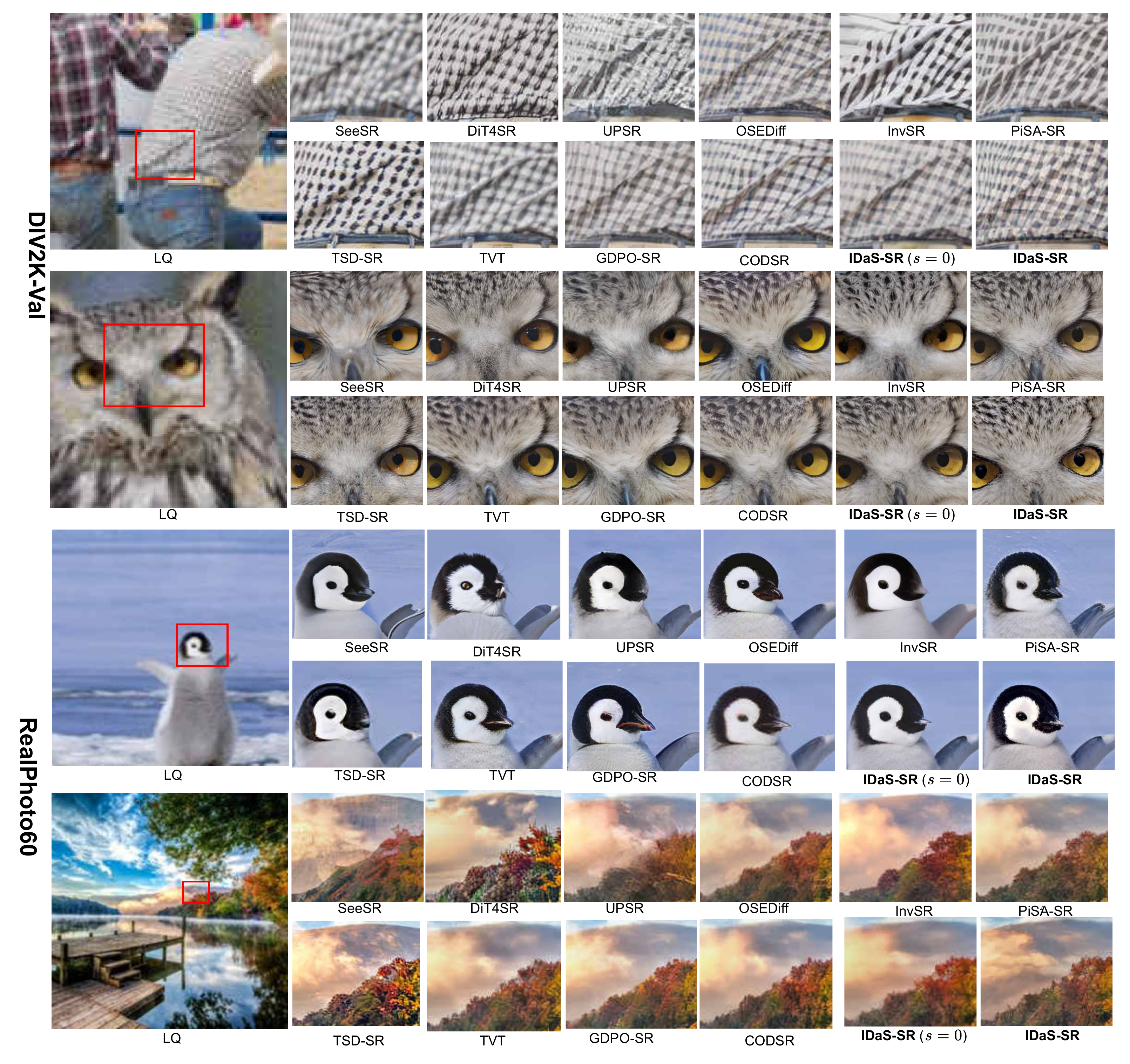}
    \caption{Qualitative comparisons on the DIV2K-Val~\cite{div2k} and RealPhoto60~\cite{supir} datasets.}
    \label{fig:visual-2}
\end{figure*}

\noindent\textbf{Anchoring timestep concentration.} 
The anchoring timestep $\hat{t}$ is an instance-adaptive structural anchor. 
Its distribution across two benchmarks (Fig.~\ref{fig:timestep_histogram}) is well-concentrated and near-Gaussian, centered at $\mu\approx115$ and bounded within 90--150, corresponding to a cumulative signal-retention factor $\bar{\alpha}\approx0.85$ (the fraction of clean-latent signal preserved at the anchor). 
Fig.~\ref{fig:timestep_histogram} contrasts this predicted anchor with the LQ's measured equivalent timestep: although the raw LQ latents sit near-clean ($t_{equiv}\!\approx\!14$--$20$, $\bar{\alpha}\!\approx\!0.93$--$0.95$, measured SNR ${\sim}11$--$13$~dB), MINE anchors them at a lower band ($\bar{\alpha}\!\approx\!0.82$--$0.91$, SNR ${\sim}6.6$--$10.0$~dB): deep enough to engage the generative prior, yet still signal-dominated to preserve structure. 
The modest spread ($\sigma\approx11$) confirms $\hat{t}$ is not fixed to a single value but varies with each input, while remaining concentrated in this restoration-suitable band.

\noindent\textbf{Degradation factorization in $\hat{c}_{deg}$.} 
While $\hat{t}$ is a scalar structural anchor, the high-dimensional $\hat{c}_{deg}$ carries observation-specific structure that a scalar cannot express. 
To probe what it encodes, we construct a balanced grid of two degradation types (blur, noise) at three severity levels over 100 shared images and remove per-image content by within-image centering, which cancels the content term exactly. 
Fig.~\ref{fig:tsne_cdeg} visualizes the resulting embeddings: $\hat{c}_{deg}$ separates degradation \emph{type} into distinct clusters while encoding \emph{severity} as a within-cluster gradient. 
This factorization, a type axis alongside a graded severity axis, is the observation-specific context that motivates conditioning on $\hat{c}_{deg}$ beyond the scalar anchor (Tab.~3 in main paper).

\section{Limitations and Future Work}
Although IDaS-SR achieves strong perceptual restoration with continuous fidelity-realism control, this capability relies on auxiliary modules (MINE and the prompt extractor) that add parameters and inference cost over minimal one-step methods (Tab.~\ref{tab:efficiency}). 
MINE anchors the low-quality input and, with CHARIOT, lets a single scalar steer continuously from faithful restoration to realistic generation, which fixed-mapping methods cannot offer. 
This cost is the trade-off for that flexibility, and improving the efficiency of these bridging modules is a promising avenue for further work.

Our analysis points to two further extensions. 
The degradation embedding $\hat{c}_{deg}$ learns degradation-relevant structure implicitly, from the restoration objective alone and without explicit labels (Sec.~\ref{sec:positioning_analysis}), suggesting that explicit degradation supervision could further strengthen this conditioning under severe or out-of-distribution corruptions. 
The single global steering scalar could likewise extend to spatially or semantically varying control, letting the fidelity-realism balance differ across an image. 
These directions offer natural paths toward a more efficient and versatile framework.

% \begin{figure}[t]
%     \centering    
%     \includegraphics[width=.9\linewidth]{Fig/timestep_distribution_realsr_drealsr_annot.pdf}
%     \caption{Predicted anchoring timestep $\hat{t}$ versus the LQ's measured equivalent timestep $t_{equiv}$ on RealSR~\cite{realsr} and DRealSR~\cite{drealsr}. The LQ latents are near-clean ($t_{equiv}\!\approx\!14$--$20$); MINE anchors them deeper at a consistent $\hat{t}\!\approx\!115$ ($\bar{\alpha}\!\approx\!0.85$, $\sigma\!\approx\!11$)---far enough to engage the generative prior while preserving structure.}
%     \label{fig:timestep_histogram}
% \end{figure}

% \begin{figure}[t]
%     \centering    
%     \includegraphics[width=\linewidth]{Fig/tsne_cond_type_severity.pdf}
%     \caption{t-SNE of $\hat{c}_{deg}$ on a balanced grid of blur/noise degradations at three severity levels, with per-image content removed by within-image centering. Type forms distinct clusters and severity within a cluster gradient. The two are independent: a type direction fitted at one severity classifies at another with 97--99\% accuracy.}
%     \label{fig:tsne_cdeg}
% \end{figure}

\begin{table*}[p]
    \centering
    \small
    \setlength{\tabcolsep}{2mm}
    \begin{tabular}{l c cc cccc}
        \toprule
        \textbf{Method} & \textbf{NFE}$\downarrow$ & \textbf{LPIPS}$\downarrow$ & \textbf{DISTS}$\downarrow$ & \textbf{CLIPIQA}$\uparrow$ & \textbf{HyperIQA}$\uparrow$ & \textbf{MUSIQ}$\uparrow$ & \textbf{TOPIQ}$\uparrow$ \\
        \midrule
        \multicolumn{8}{l}{\textit{DIV2K-Val Dataset~\cite{div2k}}} \\
        \midrule
        SeeSR               & 50 & 0.285 & 0.201 & 0.721 & \thirdbest{0.641} & \thirdbest{70.12} & \thirdbest{0.662} \\
        FaithDiff           & 20 & 0.312 & 0.218 & 0.655 & 0.598 & 68.45 & 0.591 \\
        UPSR                & 15 & \secondbest{0.262} & \best{0.188} & 0.512 & 0.525 & 58.90 & 0.505 \\
        DiT4SR              & 20 & 0.335 & 0.229 & 0.682 & 0.612 & 66.80 & 0.628 \\
        % SUPIR               & 50 & 0.385 & 0.248 & 0.640 & 0.575 & 61.80 & 0.562 \\
        \cmidrule(lr){1-8}
        SinSR               & 1  & 0.324 & 0.206 & 0.649 & 0.580 & 62.79 & 0.572 \\
        OSEDiff             & 1  & 0.278 & 0.198 & 0.718 & 0.620 & 68.90 & 0.612 \\
        InvSR               & 1  & 0.320 & 0.221 & 0.705 & 0.592 & 66.12 & 0.598 \\
        PiSA-SR             & 1  & \thirdbest{0.265} & \secondbest{0.192} & 0.712 & 0.638 & 69.80 & 0.645 \\
        TSD-SR              & 1  & 0.268 & 0.195 & \secondbest{0.742} & 0.625 & \secondbest{70.85} & 0.638 \\
        HYPIR               & 1  & 0.298 & 0.212 & 0.648 & 0.582 & 63.50 & 0.581 \\
        TVT                 & 1  & \best{0.258} & 0.194 & \thirdbest{0.735} & \secondbest{0.651} & 68.90 & \secondbest{0.671} \\
        % OMGSR               & 1  & 0.280 & 0.202 & 0.710 & 0.622 & 69.10 & 0.635 \\
        GDPO-SR             & 1  & 0.290 & 0.196 & 0.692 & 0.626 & 68.84 & 0.620 \\
        CODSR               & 1  & 0.288 & \thirdbest{0.193} & 0.694 & 0.632 & 69.59 & 0.619 \\
        % TADSR               & 1  & 0.295 & 0.215 & 0.745 & 0.662 & 70.80 & 0.680 \\
        \cmidrule(lr){1-8}
        IDaS-SR             & 1  & 0.298 & 0.212 & \best{0.755} & \best{0.668} & \best{71.20} & \best{0.695} \\
        \midrule
        \multicolumn{8}{l}{\textit{DRealSR Dataset~\cite{drealsr}}} \\
        \midrule
        SeeSR               & 50 & 0.314 & 0.230 & 0.690 & \thirdbest{0.658} & 64.75 & \thirdbest{0.653} \\
        FaithDiff           & 20 & 0.350 & 0.238 & 0.634 & 0.612 & 66.07 & 0.575 \\
        UPSR                & 15 & \secondbest{0.286} & \best{0.205} & 0.483 & 0.502 & 56.27 & 0.484 \\
        DiT4SR              & 20 & 0.371 & 0.245 & 0.660 & 0.604 & 64.60 & 0.648 \\
        % SUPIR               & 50 & 0.443 & 0.281 & 0.625 & 0.531 & 53.71 & - \\
        \cmidrule(lr){1-8}
        SinSR               & 1  & 0.369 & 0.249 & 0.643 & 0.516 & 55.54 & 0.513 \\
        OSEDiff             & 1  & 0.297 & 0.216 & 0.696 & 0.605 & 64.69 & 0.590 \\
        InvSR               & 1  & 0.354 & 0.246 & 0.713 & 0.615 & 66.00 & 0.611 \\
        PiSA-SR             & 1  & 0.296 & 0.217 & 0.697 & 0.639 & 66.11 & 0.633 \\
        TSD-SR              & 1  & 0.297 & \thirdbest{0.213} & \secondbest{0.735} & 0.613 & \thirdbest{66.60} & 0.618 \\
        HYPIR               & 1  & 0.329 & 0.231 & 0.627 & 0.577 & 61.47 & 0.569 \\
        TVT                 & 1  & \thirdbest{0.290} & 0.221 & \thirdbest{0.722} & \best{0.666} & 65.56 & \secondbest{0.659} \\
        % OMGSR               & 1  & 0.327 & 0.233 & 0.696 & 0.620 & 66.73 & 0.625 \\
        GDPO-SR             & 1  & \best{0.268} & \secondbest{0.211} & 0.703 & 0.616 & 65.59 & 0.603 \\
        CODSR               & 1  & 0.292 & \secondbest{0.211} & 0.715 & 0.623 & \best{67.03} & 0.606 \\
        % TADSR               & 1  & 0.323 & 0.241 & 0.740 & 0.664 & 67.01 & - \\
        \cmidrule(lr){1-8}
        IDaS-SR             & 1  & 0.329 & 0.239 & \best{0.746} & \secondbest{0.659} & \secondbest{66.81} & \best{0.674} \\
        \midrule
        \multicolumn{8}{l}{\textit{RealSR Dataset~\cite{realsr}}} \\
        \midrule
        SeeSR               & 50 & 0.300 & 0.222 & 0.667 & \thirdbest{0.672} & 69.70 & \secondbest{0.686} \\
        FaithDiff           & 20 & 0.288 & 0.211 & 0.615 & 0.622 & 68.77 & 0.615 \\
        UPSR                & 15 & 0.286 & 0.208 & 0.529 & 0.565 & 65.06 & 0.571 \\
        DiT4SR              & 20 & 0.316 & 0.223 & 0.628 & 0.613 & 67.96 & 0.601 \\
        % SUPIR               & 50 & 0.412 & 0.265 & 0.633 & 0.580 & 63.20 & 0.585 \\
        \cmidrule(lr){1-8}
        SinSR               & 1  & 0.317 & 0.235 & 0.621 & 0.527 & 60.82 & 0.520 \\
        OSEDiff             & 1  & 0.292 & 0.213 & 0.669 & 0.608 & 69.08 & 0.597 \\
        InvSR               & 1  & 0.287 & 0.212 & 0.679 & 0.528 & 68.54 & 0.590 \\
        PiSA-SR             & 1  & \secondbest{0.267} & \thirdbest{0.204} & 0.670 & 0.641 & 70.15 & 0.637 \\
        TSD-SR              & 1  & 0.272 & 0.209 & \secondbest{0.693} & 0.642 & \best{70.72} & 0.658 \\
        HYPIR               & 1  & 0.305 & 0.227 & 0.639 & 0.600 & 66.25 & 0.598 \\
        TVT                 & 1  & \best{0.260} & 0.206 & 0.689 & \secondbest{0.676} & 69.89 & \thirdbest{0.683} \\
        % OMGSR               & 1  & 0.295 & 0.215 & 0.675 & 0.625 & 69.80 & 0.632 \\
        GDPO-SR             & 1  & \thirdbest{0.268} & \best{0.198} & 0.677 & 0.617 & 69.07 & 0.608 \\
        CODSR               & 1  & 0.274 & \secondbest{0.200} & \thirdbest{0.690} & 0.632 & \secondbest{70.51} & 0.609 \\
        % TADSR               & 1  & 0.317 & 0.232 & 0.728 & 0.688 & 71.19 & - \\
        \cmidrule(lr){1-8}
        IDaS-SR             & 1  & 0.325 & 0.236 & \best{0.734} & \best{0.680} & \thirdbest{70.49} & \best{0.711} \\
        \bottomrule
    \end{tabular}
    \caption{Complete quantitative comparison on DIV2K-Val~\cite{div2k}, DRealSR~\cite{drealsr}, and RealSR~\cite{realsr}. IDaS-SR is reported at the default generative steering setting ($s{=}0.6$). Best, second, and third are \best{bolded}, \secondbest{underlined}, and \thirdbest{italicized}, respectively.}
    \label{tab:extended_full}
\end{table*}

% {
%     \small
%     \bibliographystyle{ieeenat_fullname}
%     \bibliography{main}
% }

\end{document}